\documentclass{ieeeaccess}
\usepackage{cite}
\usepackage{amsmath,amssymb,amsfonts}
\usepackage{algorithmic}
\usepackage{graphicx}
\usepackage{multirow}  
\usepackage{textcomp}
\usepackage{subfigure}
\usepackage{url}
\usepackage{booktabs}

\def\thevol{X}

\def\BibTeX{{\rm B\kern-.05em{\sc i\kern-.025em b}\kern-.08em
    T\kern-.1667em\lower.7ex\hbox{E}\kern-.125emX}}

\begin{document}
\title{Privacy Meets Explainability: \\A Comprehensive Impact Benchmark}

\author{\uppercase{Saifullah Saifullah}\authorrefmark{1,2}*, \uppercase{Dominique Mercier}\authorrefmark{1,2}*, \uppercase{Adriano Lucieri}\authorrefmark{1,2}*
\uppercase{Andreas Dengel\authorrefmark{1,2}, and \uppercase{Sheraz Ahmed}}\authorrefmark{2}}
\address[1]{Department of Computer Science, Technische Universit\"at Kaiserslautern, Erwin-Schrödinger-Straße 52, 67663 Kaiserslautern, Germany}
\address[2]{Smart Data and Knowledge Services (SDS), DFKI GmbH, 67663 Kaiserslautern, Germany (e-mail: firstname.lastname@dfki.de)}
\tfootnote{*Equal contribution \\ This work was supported by the BMBF projects SensAI (BMBF Grant 01IW20007) and ExplAINN (BMBF Grant 01IS19074). We thank all members of the Deep Learning Competence Center at the DFKI for their comments and support.}

\markboth
{Saifullah \headeretal: Privacy Meets Explainability: A Comprehensive Impact Benchmark}
{Saifullah \headeretal: Privacy Meets Explainability: A Comprehensive Impact Benchmark}

\corresp{Corresponding author: Saifullah Saifullah (E-mail: saifullah.saifullah@dfki.de).}

\begin{abstract}
Since the mid-10s, the era of Deep Learning (DL) has continued to this day, bringing forth new superlatives and innovations each year.
Nevertheless, the speed with which these innovations translate into real applications lags behind this fast pace.
Safety-critical applications, in particular, underlie strict regulatory and ethical requirements which need to be taken care of and are still active areas of debate.
eXplainable AI (XAI) and privacy-preserving machine learning (PPML) are both crucial research fields, aiming at mitigating some of the drawbacks of prevailing data-hungry black-box models in DL.
Despite brisk research activity in the respective fields, no attention has yet been paid to their interaction.
This work is the first to investigate the impact of private learning techniques on generated explanations for DL-based models.
In an extensive experimental analysis covering various image and time series datasets from multiple domains, as well as varying privacy techniques, XAI methods, and model architectures, the effects of private training on generated explanations are studied.
The findings suggest non-negligible changes in explanations through the introduction of privacy.
Apart from reporting individual effects of PPML on XAI, the paper gives clear recommendations for the choice of techniques in real applications.
By unveiling the interdependencies of these pivotal technologies, this work is a first step towards overcoming the remaining hurdles for practically applicable AI in safety-critical domains.
\end{abstract}


\titlepgskip=-15pt

\maketitle

\section{Introduction}
\PARstart{I}{n} recent years, a wide variety of deep learning (DL) approaches have achieved outstanding performance in a wide range of application domains~\cite{liu2019comparison,sujatha2021performance}. The versatile applications of deep neural networks in areas such as image processing~\cite{hemanth2017deep}, object segmentation~\cite{chen2013deep}, document analysis~\cite{gilani2017table}, time series classification~\cite{ismail2019deep}, time series prediction~\cite{lim2021time}, layout classification~\cite{binmakhashen2019document}, sensor analysis and other areas have contributed to immense growth.
Intelligent and automated decision-making bears the potential to improve and transform a row of critical application domains, including finance, healthcare, transportation, and administration. However, DL-based systems rely on complex, data-driven black-box methods whose exact working mechanisms are still widely unexplained in the scientific community, while the secure application of algorithms in safety-critical domains requires transparency and traceability of decisions. Furthermore, data security is a serious concern in domains involving critical and personal data. DL methods are data-driven, often requiring the transmission, processing, and storage of large amounts of data in multiple remote locations. A variety of works showed, that even after training, neural networks can leak sensitive information about training data~\cite{fredrikson2015model,chen2021understanding}. The lack of explainability and privacy of modern DL systems are some of the main challenges that prevent the practical use of these powerful methods in safety-critical domains.

The field of eXplainable AI (XAI) seeks to unveil the decision-making processes of black-box models and has been thoroughly researched in recent years~\cite{das2020opportunities}. Depending on the area of application, a variety of different methods have been developed that attempt to explain the prediction of networks as well as their underlying decision-making process. Especially in image processing, so-called attribution methods are often used~\cite{nielsen2021robust}. These methods generate heatmaps that highlight the areas of the input that were significantly involved in the network prediction. Therefore, these methods may or may not need access to the networks' internals. While this type of explainability is widely used, it is not always adequate. In some cases, an explanation for the complete process is required. This is beyond the capabilities of attribution methods, which provide local explanations.
One way of approaching global explanations is the training of architectures that yield explanations by design. Well-known representatives of this group are prototype-based systems~\cite{li2018deep}. However, the special requirements regarding the model architecture introduce additional limitations. Although it is arguable whether any XAI method allows explaining a given process in its entirety, at least the representation learned by such a network is more interpretable and related to known concepts, as opposed to networks without an intrinsic interpretable design.

Some properties of DL-based models (i.e., gradients) are of great importance for decision explanation. However, these properties also provide interfaces for the targeted retrieval of sensitive information. It has been shown that only limited model access suffices to completely reconstruct models and steal their training data~\cite{fredrikson2015model}. This constitutes a major risk for the deployment of AI in safety-critical applications. Moreover, it is a significant threat to individuals contributing to a model's training data and users alike~\cite{liu2020privacy}. To prevent this, many methods have been designed to protect neural networks from attacks during the training process and to reduce data leakage during later deployment. These methods also open new opportunities for collaboration between multiple parties, allowing for better predictions based on larger amounts of data.

While XAI methods increase the transparency and intelligibility of a model's decision-making behavior, privacy-protection techniques prevent the leakage of sensitive information. However, the safe deployment of data-driven systems in safety-critical areas is only possible if one can reconcile both goals.
So far, there has been no work that investigates and describes the specific impact of different privacy-preserving methods on the quality of explanations in deep learning.
Understanding the trade-off between both concurring objectives is crucial to improve XAI methods and assure their correct interpretation in a privacy-preserving setting, constituting an important step in the practical applicability of DL.

This work is the first to thoroughly analyze the influence of different privacy-preserving machine learning (PPML) techniques on the explanations generated by XAI methods. 
In an extensive, multivariate analysis, three different privacy techniques (\textit{Differential Privacy} (\textit{DP})~\cite{abadi2016deep}, \textit{Federated Learning} (\textit{FedAVG})~\cite{konevcny2016federated}, and \textit{Differential Private Federated Learning} (\textit{FedAVG-DP})) are combined with a series of XAI attribution methods, and applied to seven different model architectures trained on eleven different datasets from various domains including document image, natural image, medical image, as well as time series analysis.
The evaluation qualitatively and quantitatively highlights the varying but non-negligible impact of PPML methods on the quality of explanations.

The analysis reveals important relationships between private training and XAI.
\begin{itemize}
    \item \textit{Differential Privacy} hampers the \textit{Interpretability} of explanations.
    \item \textit{Federated Learning} often facilitates the interpretation of generated explanations.
    \item The \textit{Fidelity} of explanations is potentially deteriorated when using \textit{DP}.
    \item The negative effects introduced by \textit{DP} can be moderated by combining it with \textit{FedAVG}.
    \item Perturbation-based XAI methods are less affected by \textit{DP}-based training procedures.
\end{itemize}

The remainder of this paper is structured as follows. 
Section~\ref{sec:related_work} gives an overview of relevant XAI methods and PPML techniques. 
The various datasets used throughout this study are introduced in Section~\ref{sec:datasets}. 
In  Section~\ref{sec:experiments_and_results} the complete experimental setup is outlined, followed by the presentation of the respective results. 
Trends and findings of this analysis are discussed throughout Section~\ref{sec:discussion} and the manuscript is concluded in Section~\ref{sec:conclusion}.

\begin{table*}[!t]
\renewcommand{\arraystretch}{1.3}
\caption{Shows the datasets used to evaluate the impact of PPML on XAI methods. The datasets cover the image, document image, medical image and time series classification.}
\label{tab:datasets}
\centering
\begin{tabular}{llrrrrr}
\toprule
\textbf{Modality \& Dataset} & \textbf{Domain} & \textbf{Train} & \textbf{Test} & \textbf{Dimensions} & \textbf{Channels} & \textbf{Classes} \\
\midrule
\textbf{Time Series} & & & & & \\
Anomaly Detection       & Synthetic     & 50,000    & 10,000    & 50    & 3 &2 \\
Character Trajectories  & Communication & 1,422     & 1,436     & 182   & 3 & 20 \\
ECG5000                 & Medical       & 500       & 4,500     & 140   & 1 & 5 \\
FordA                   & Manufacturing & 3,601     & 1,320     & 500   & 1 & 2 \\
Wafer                   & Information   & 1,000     & 6,164     & 152   & 1 & 2 \\
\midrule
\textbf{Images} & & & & & \\
RAF-Database            & Facial Expressions Recognition        & 12,271     & 3,068    & $224 \times 224$    & 3 & 7 \\
Caltech-256             & Natural Image Classification & 24,485    & 6,122    & $224 \times 224$    & 3 & 256 \\
ISIC                    & Medical Image Analysis    & 26,521     & 2,947    & $224 \times 224$   & 3 & 8 \\
SCDB                    & Synthetic     & 6,000     & 1,500     & $224 \times 224$   & 3 & 2 \\
RVL-CDIP                & Document Analysis & 320,000    & 40,000    & $224 \times 224$    & 3 & 16 \\
Tobacco3482             & Document Analysis & 2,782     & 700     & $224 \times 224$    & 3 & 10 \\
\bottomrule
\end{tabular}
\end{table*}

\section{Related Work}
\label{sec:related_work}
In the following section, the most common methods in the fields of XAI and PPML will be briefly outlined.
A full review of methods is beyond the scope of this work.
The interested reader can find extensive reviews of XAI and PPML in \cite{vilone2020explainable} and \cite{boulemtafes2020review}, respectively.

\subsection{XAI}
In the field of XAI, attribution methods are very widely used due to their versatility and comprehensibility.
Attribution maps approximate the relevance of input features or feature groups to the local model decision and belong to the group of so-called \textit{post-hoc methods}~\cite{fan2021interpretability}, which are mainly characterized by their ability to explain models that have already been trained.
In 2013, the \textit{Saliency}~\cite{simonyan2013deep} was published as one of the first methods in this field based on the backpropagation~\cite{lecun2015deep} algorithm used to train networks.
An extension of this method is \textit{InputXGradient}~\cite{shrikumar2016not}, in which the coherence of input features is additionally taken into account. 
Other methods that work similarly to the aforementioned methods include \textit{GuidedBackpropagation}~\cite{springenberg2014striving} and \textit{IntegratedGradients}~\cite{sundararajan2017axiomatic}. 
All these methods are so-called \textit{gradient-based methods} and need access to the networks' internals to compute explanations. 

In contrast, \textit{perturbation-based} methods are usually \textit{model-agnostic}, therefore not requiring any specific model architecture to work on.
The \textit{Occlusion}~\cite{zeiler2013visualizing} method removes input areas sequentially and reevaluates each manipulated input to measure the influence of single regions on the network prediction.
Another subgroup of \textit{perturbation-based} algorithms derives surrogate models from the local model behavior.
\textit{Local Interpretable Model-Agnostic Explanations} (\textit{LIME})~\cite{ribeiro2016should}, for instance, applies perturbations to an input sample to obtain a local linear model from these inputs and the respective model predictions.
\textit{Shapley Additive Explanations} (\textit{SHAP})~\cite{lundberg2017unified} has been proposed as a related method, with additional constraints based on game theory to provide certain mathematical guarantees. 

For the sake of completeness, it should be mentioned that there are several other approaches besides the discussed \textit{post-hoc} attribution methods. 
\textit{Prototype-based}~\cite{li2018deep}, \textit{patch-based}~\cite{chen2019looks}, and \textit{concept-based} methods~\cite{kim2018interpretability}, for instance, have also been used previously. 

\subsection{PPML}
Different techniques have been developed to protect data and model privacy during the training and in the subsequent inference phase. 
Anonymisation techniques (e. g., K-anonymity~\cite{hellani2015towards}) were among the first approaches developed to ensure privacy in model training.
In the meantime, there have been outstanding breakthroughs in the area of privacy attacks. 
Membership~\cite{rahman2018membership} or model inversion attacks~\cite{fredrikson2015model} allow reconstructing training data with extremely limited access to the models.
Therefore, simplistic techniques such as anonymization are no longer sufficient. 

A promising training technique that leads to a high degree of privacy for data and model is \textit{Homomorphic Encryption}~\cite{aono2017privacy}. 
However, this method is rarely applicable to modern DL-based systems, due to its massive computational overhead.
Another, more frequently used technique is \textit{Differential Privacy} (\textit{DP})~\cite{abadi2016deep}. 
Here, a certain amount of noise is added to the training signal of deep networks to prevent its parameters to capture information held by specific training samples but instead focus on the general characteristics of the whole population. 
One advantage of this method is that it can be applied to a wide variety of architectures and requires only minimal changes to the training setup. 
Another prominent technique used to account for data-privacy is \textit{Federated Learning} (\textit{FedAVG})~\cite{konevcny2016federated}.
In \textit{FedAVG}, local models are trained on a data-owners subset of training samples, and only the locally computed gradients are sent to a centralized server. 
There, the average is calculated to obtain a global model.
This way, sensitive data does not need to leave the institution, but multiple institutions can collaborate to leverage a bigger training set for the global model.
Moreover, \textit{FedAVG} can be combined with \textit{DP} to prevent the risk of data leakage from model gradients. 
Out of the many other privacy techniques~\cite{liu2020privacy}, \textit{DP} and \textit{FedAVG} stand out as the most commonly used.

To this date, both XAI and PPML have mostly been considered in separation by the research community. 
Few works~\cite{rahman2021secure, franco2021toward} have made the first attempts at combining explanation and privacy preservation in a single framework. 
However, there is still a lack of insight regarding the exact influence private training has on the generated explanations in deep learning. 

\section{Datasets}
\label{sec:datasets}

To comprehensively analyze the impact of privacy-preserving methods on explanations, a variety of different datasets from different domains in time series and image analysis were utilized, as listed in Table~\ref{tab:datasets}. 

\subsection{Time series Datasets}
The first modality that is evaluated uses time series data. Time series data is usually acquired using different sensors and differs from image data concerning various characteristics such as the locality constraints and the dependence on a sequential order. 

With the exception of the \textit{Anomaly Detection} dataset~\cite{siddiqui2019tsviz}, the datasets for the time series analysis come from the UEA \& UCR repository~\cite{tsc2021datasets}. This selection includes both univariate and multivariate time series with different numbers of classes. The \textit{Anomaly Detection} dataset and the \textit{FordA} dataset consider the task of anomaly detection. The \textit{Anomaly Detection} dataset deals with point anomalies and the \textit{FordA} dataset with sequence anomalies. The point anomalies are very interpretable for humans as in their case the data is more or less noise and contains a large peak that indicates the anomaly. Even without the annotation, it is possible to understand whether the explanation for such a sample is correct or not. This is not the case for the \textit{FordA} data as the sequences are very long and there is no annotation. In this dataset, the anomaly can be a long part of the sequence that varies from the expected behavior. The \textit{Character Trajectories} dataset was selected as it is possible to transform it back to the 2d input space to understand the explanation. It consists of three channels covering the acceleration within the x and y direction and the pen force. Therefore, it is a real-world dataset that enables precise identification of whether an explanation is good or not. In addition, it is important to mention that the dataset size of the time series datasets differs significantly, to properly represent the influence of data volume.

\subsection{Image Datasets}
\subsubsection{Natural Image Datasets}
Image datasets range from specialized domains such as facial recognition, medical image analysis, and document analysis to toy datasets covering a varying number of classes, channels, and dataset sizes.
The Real-world Affective Faces Database (\textit{RAF-Database})~\cite{li2019reliable} is a collection of 15,339 face images with crowd-sourced annotations. It classifies images in one of the seven facial expressions (Surprise, Fear, Disgust, Happiness, Sadness, Anger, Neutral). \textit{Caltech-256}~\cite{griffin2007caltech} is a natural images classification dataset comprising of a total of 30,607 labeled images with 256 unique object classes.
\subsubsection{Medical Image Datasets}
The International Skin Imaging Collaboration (ISIC) provides a large publicly accessible library of digital skin images\footnote{The ISIC Archive is accessible at \url{https://www.isic-archive.com}.} and hosts annual challenges. The \textit{\textit{ISIC}} dataset used in this work is a cleaned combination of all ISIC challenge datasets. The datasets have been merged and freed from duplicates according to the recommendations in~\cite{cassidy2022analysis}. All images are labeled as either Melanoma (\textit{MEL}), Nevus (\textit{NV}), Basal Cell Carcinoma (\textit{BCC}), Actinic Keratosis (\textit{AK}), Benign Keratotic Lesion (\textit{BKL}), Dermatofibroma (\textit{DF}), Vascular Lesion (\textit{VASC}) or Squamous Cell Carcinoma (\textit{SCC}). The classification is based on complex combinations of distributed and overlapping biomarkers, posing particular challenges for the explanation of automated decisions.
The seven-point checklist criteria dataset (\textit{Derm7pt}) proposed in~\cite{kawahara2018seven} consists of clinical and dermoscopic images of 1.011 skin lesions with extensive annotation.
In this work, only the subset of dermoscopic images along with the respective diagnosis annotations for pre-training of ISIC classifiers is considered in experiments involving \textit{DP}.
The \textit{SCDB}~\cite{lucieri2020explaining} dataset is a synthetic toy dataset inspired by the problems of skin lesion analysis. Images are classified into one of two classes based on the combinations of shapes present in a base shape, depicting the skin lesion. The shapes can be overlapping and redundant, but classification evidence is sparse and localized. Along with the class label, each image is supplemented by shape annotation maps, serving as ground truth explanations.

\subsubsection{Document Image Datasets}
Business documents are a fundamental component of modern industry. Recent advances in deep learning have sparked a growing interest in automating document processing tasks such as document search, and extraction of document information. However, business documents often contain highly personal user data and sensitive information pertaining to a company's intellectual property, which makes the secure application of Deep Learning in this area a major concern. On the other hand, deep learning-based decision-making processes have been shown to be susceptible to learning biases in the data \cite{Ntoutsi-deep-learning-bias}. An example of such a system involves automatically analyzing resumes to make hiring decisions, which may lead to discrimination against women or members of minority groups. Explainability of such systems is therefore of paramount importance for their safe and practical deployment.

In order to analyze the interdependence between PPML and XAI for document domain, two popular document benchmark datasets are utilized in this study. \textit{RVL-CDIP}~\cite{harley2015icdar} is a large-scale document dataset that has been extensively used as a benchmark for document analysis tasks. The dataset contains a total of 400,000 labeled document images with 16	different categories and consists of training, testing, and validation split of 320,000, 40,000, and 40,000 images respectively. 
\textit{Tobacco3482}\footnote{\url{https://www.kaggle.com/patrickaudriaz/tobacco3482jpg}.}, is another popular but small-scale dataset with 3482 labeled document images. Since there is no split defined for this dataset, the training and test, and validation splits of sizes 2,504, 700, and 278 images were defined. Since both of these datasets are subsets of a bigger dataset, there exists some overlap between them. Therefore, for all the experiments, the overlapping images were removed from the \textit{RVL-CDIP} dataset. The two datasets were used in combination to analyze the effects of transfer learning on the privacy and explainability aspects of the models.

\section{Experiments \& Results}
\label{sec:experiments_and_results}
A broad experimental basis, covering various domains, applications, and configurations is necessary, to make general statements about the impact of privacy techniques on explanations.
Therefore, a selection of state-of-the-art classifiers is trained on a range of different datasets and applications covering both time series and image domains.
Each combination of model and dataset is trained in four different settings, including training without privacy (\textit{Baseline}), with differential privacy (\textit{DP}), federated training (\textit{FedAVG}) and federated training with client-side differential privacy (\textit{FedAVG-DP}).
Different explanation methods are finally applied to every model instance to compare their generated explanations.

Evaluating explanations and judging their quality is a common problem not only in XAI research~\cite{zhou2021evaluating}, but also in the social sciences~\cite{miller2019explanation}.
Multiple evaluation dimensions have to be considered to make clear statements about the impact of privacy-preserving model training on the explainability of DL-based models.
Human-centered evaluation is laborious and requires domain experts.
Instead, functionality-grounded methods are best suited for the domain- and dataset-wide fair comparison and quality assessment of XAI and are therefore utilized throughout this study.

In the experiments, the focus lies on the two main properties of explanations as defined in~\cite{zhou2021evaluating}, namely their \textbf{Fidelity} and \textbf{Interpretability}.
\textit{Fidelity} measures soundness and completeness to ensure that explanations accurately reflect a model's decision-making behavior.
\textit{Interpretability} refers to the clarity, parsimony, and broadness of explanations, and therefore describes factors related to the ease of communication on the interface of machines and humans. Functionality-grounded methods make use of formal mathematical definitions as proxies of perceived interpretability.

In this section, details of the experimental setup are described, and evaluation metrics are introduced. Afterward, the results and analysis of all setups are thoroughly presented.

\subsection{Experiment Setup}

Baseline networks were trained using the standard SGD or ADAM optimizers with varying numbers of epochs per dataset, to ensure convergence. 
For all other settings, training and privacy hyperparameters have been manually tuned to find a good trade-off between privacy and model performance matching the baseline.
This is important to guarantee a sufficiently fair comparison between the methods since a significantly worse network would also show worse attribution results. 
However, all models were trained with overall comparable settings. 
Moreover, fixed seeds were used to ensure reproducibility. 
The reported performances correspond to the accuracies of the model performing best on the test datasets.
In time series analysis, \textit{InceptionTime}~\cite{ismail2020inceptiontime} and \textit{ResNet-50}~\cite{feng2017overview} were used as representative networks, since they achieve state-of-the-art performances for the utilized datasets. 
\textit{ResNet-50}, \textit{NFNet}~\cite{brock2021high} and \textit{ConvNeXt}~\cite{liu2022convnet} have been used for classification of natural and medical images. 
Since document images differ significantly from natural images, a different set of models has been used for this domain, including \textit{AlexNet}~\cite{krizhevsky2017imagenet}, \textit{VGG-16}~\cite{simonyan2014very}, \textit{ResNet-50}, \textit{EfficientNet}~\cite{tan2019efficientnet}, and \textit{ConvNext}, which have shown the best performance in the past.
The training data was split between training and validation with a factor of 0.9, wherever no validation dataset had been provided. 

Some XAI methods pose specific requirements on the model architecture or training procedure, complicating the application of privacy-protection techniques. 
Therefore, this work solely focus on the commonly used \textit{post-hoc} explanations.
Different attribution methods vary considerably in their realization and their associated underlying assumptions.
Therefore, it was decided to apply a broad range of diverse methods differing in their implementations and theoretical foundations.
The work covers a total of nine XAI methods, including gradient-based Saliency~\cite{simonyan2013deep}, InputXGradient~\cite{shrikumar2016not}, GuidedBackpropagation~\cite{springenberg2014striving}, IntegratedGradients~\cite{sundararajan2017axiomatic}, DeepSHAP~\cite{lundberg2017unified}, and DeepLift~\cite{shrikumar2017learning}, but also gradient-free methods such as Occlusion~\cite{zeiler2013visualizing}, LIME~\cite{ribeiro2016should} and KernelSHAP~\cite{lundberg2017unified}.

\subsection{Evaluation Metrics}
The \textit{Fidelity} of the explanations is quantified using \textit{Sensitivity}~\cite{yeh2019fidelity}, \textit{Infidelity}~\cite{yeh2019fidelity}, \textit{Area Over the Perturbation Curve}~\cite{samek2016evaluating}, and \textit{Ground Truth Concordance} evaluation metrics.
When measuring the \textit{Sensitivity}, insignificant perturbations are applied to the input and the change in the attribution map is measured. Small changes in the input should not result in large changes in the attribution map. Thus, a smaller value is better. 
\textit{Infidelity} applies significant perturbations to the input and the corresponding attribution map. The value measures the mean-squared error between the perturbed heatmap and the difference in the predictions of perturbed and unperturbed input.
The \textit{Area Over the Perturbation Curve} (\textit{AOPC}) measures the alignment between an attribution map's relevance values and the effect of perturbing the corresponding input regions on the model prediction. Intuitively, removing features with lower importance should affect the prediction less than the deletion of important features. The \textit{AOPC} is a scalar value computed by integrating the curve over the impact of consecutive perturbations with decreasing attribution importance, relative to random perturbations. Therefore, higher values indicate an attribution map's meaningfulness.
Most fidelity measures evaluate the degree to which an explanation is faithful to the local model behavior. Whether the explanation is human-aligned, on the other hand, can only be evaluated with ground truth explanations available. The \textit{Ground Truth Concordance} was measured by computing the overlap between the highest attributed input region with the segmentation maps of the ground truth explanations. The real-valued attribution maps are binarized to allow comparison with the binary segmentations. The overlap is computed for increasing binarization thresholds from zero to one. The overlap is quantified
through the \textit{Intersection over Union} (\textit{IoU}).

The \textit{Interpretability} of explanations was measured using the \textit{Continuity} metric.
In general, humans have difficulties interpreting information that is both high dimensional and scattered.
\textit{Continuity} is defined as the sum of the absolute changes between two consecutive importance scores in an attribution map. 
For time series, the continuity is the absolute change between each subsequent point in a sequence whereas in the image domain, the absolute changes are measured and aggregated separately in X and Y directions. Better \textit{Interpretability} is indicated by lower continuity scores.

All evaluation metrics were computed on the respective test datasets. The attribution maps were normalized to have zero mean and unit standard deviation for a fair comparison across methods and different privacy types. Due to computational and time restrictions, the influence of PPML on attribution methods was quantified using a subset of the respective test sets, limited to a maximum of 1,000 examples. This work assumes that 1,000 randomly selected examples represent a sufficient quantity to generalize the findings to the complete test datasets.

\subsection{Critical Difference Diagrams}
Intuitive visualization of high-dimensional data is particularly challenging when the data origins from multiple distinct configurations as in this case.
Critical Difference (CD) diagrams, proposed by Dem{\v{s}}ar in 2006~\cite{demsar2006statistical}, allow the high-level visualization of complex experimental data intuitively and were therefore chosen to present most quantitative results.
Their ability to condense ordinal information across different datasets, models and attribution methods extracts relevant information and helps to pick up universal trends in a benchmark study.
Moreover, the method includes statistical tests, indicating the data's significance.

CD diagrams report the average rank of a given item, in a series of different settings.
If the statistical significance for two distinct items is not guaranteed, these items are connected by a horizontal line and referred to as a "clique".
In this study, the Friedman test is used to decide the statistical significance of a group of different observations.
The Holm-adjusted Wilcoxon's signed rank test is then applied for post-hoc analysis as suggested in~\cite{benavoli2016should}.
For all statistical analyses, an $\alpha$ of $0.05$ is assumed.

\subsection{Impact on Model Performance}

\begin{table}[!t]
	\renewcommand{\arraystretch}{1.3}
	\caption{Test accuracies on all datasets for different architectures and privacy-preserving settings, divided by application domain. For configurations containing \textit{DP} or \textit{FedAVG}, the $\epsilon$ and $n_c$ values are provided, respectively.} 
	\label{tab:acc}
	\centering
	\scalebox{0.75}{
	\begin{tabular}{llcccc}
		\toprule
		 & \parbox{2cm}{\textbf{Datasets \&\\Models}} & \textbf{Acc\textsubscript{Baseline}}& \textbf{Acc\textsubscript{DP} / $\epsilon$}& \textbf{Acc\textsubscript{FedAVG}} & \textbf{Acc\textsubscript{FedAVG-DP} / $\epsilon$}\\
		 \midrule
		 \multirow{19}{*}{\rotatebox{90}{\textbf{Time Series Analysis}}} & \textbf{Anomaly} &&& $n_c=4$ & \parbox{2cm}{\centering$n_c=4$}\\
		& InceptionTime	                &   98.74 & 92.87 / 5.0 & 98.77 & 89.50 / 5.0 \\
		& ResNet-50	                &   98.70 &	97.02 / 5.0 &	98.60 &	97.36 / 5.0 \\
		\cmidrule{2-6}
   	    & \textbf{Character Trajectories} &&& $n_c=4$ & \parbox{2cm}{\centering$n_c=4$}\\
		& InceptionTime	                &   99.44 &	91.85 / 5.0 &	98.82 &	87.26 / 50.0 \\
		&             	                &         &	            &	      &	68.73 / 5.0 \\
		& ResNet-50	                &   99.44 &	85.03 / 5.0 &	98.19 &	82.10 / 50.0 \\
		&             	                &         &	            &	      &	59.19 / 5.0 \\
		\cmidrule{2-6}
		& \textbf{ECG5000} &&& $n_c=4$ & \parbox{2cm}{\centering$n_c=4$}\\
		& InceptionTime	                &   94.38 &	89.07 / 5.0 &	93.36 & 89.29 / 5.0 \\
		& ResNet-50	                &   94.16 &	89.64 / 5.0 &	92.78 &	88.87 / 5.0 \\
		\cmidrule{2-6}
		& \textbf{FordA} &&& $n_c=4$ & \parbox{2cm}{\centering$n_c=4$}\\
		& InceptionTime	                &   95.61 &	92.88 / 5.0 &	97.70 &	94.17 / 50.0 \\
		&             	                &         &	            &	      &	91.43 / 5.0 \\
		& ResNet-50	                &   94.32 &	86.14 / 5.0 &	93.94 &	87.12 / 50.0 \\
		&             	                &         &	            &	      &	76.44 / 5.0 \\
		\cmidrule{2-6}
		& \textbf{Wafer} &&& $n_c=4$ & \parbox{2cm}{\centering$n_c=4$}\\
		& InceptionTime	                &   99.22 &	89.21 / 5.0 &	97.81 & 89.21 / 5.0 \\
		& ResNet-50	                &   98.75 &	89.21 / 5.0 &	89.21 & 89.21 / 5.0 \\
		\midrule
		\multirow{12}{*}{\rotatebox{90}{\textbf{Document Analysis}}} & \textbf{RVL-CDIP} &&& $n_c=8$ & \parbox{2cm}{\centering$n_c=8$}\\
		& AlexNet	                &   87.90 & 70.30 / 4.5 & 85.54 & 61.35 / 5.3 \\
		& VGG-16	                &   91.00 &	69.67 / 4.4 & 89.41 & 62.38 / 5.4 \\
		& ResNet-50	            &   90.50 &	72.55 / 5.0 & 88.25 & 68.85 / 8.8 \\
		& Efficientnet-B4	        &   92.60 &	60.20 / 4.2 & 90.59 & 45.09 / 6.5 \\
		& ConvNeXt-B	            &   93.64 &	75.60 / 3.7 & 92.60 & 73.23 / 7.7 \\
		\cmidrule{2-6}
   	    & \textbf{Tobacco3482} &&& $n_c=4$ & \parbox{2cm}{\centering$n_c=4$}\\
		& AlexNet	                &   89.57 & 86.14 / 3.9 & 91.85 & 85.71 / 8.0 \\
		& VGG-16	                &   94.14 &	85.14 / 4.9 & 93.99 & 87.00 / 7.5 \\
		& ResNet-50	            &   92.57 &	75.42 / 2.7 & 92.14 & 78.43 / 7.3\\
		& Efficientnet-B4	        &   94.42 &	89.42 / 4.4 & 93.99 & 88.57 / 8.0 \\
		& ConvNeXt-B	            &   94.71 &	87.14 / 4.8 & 94.85 & 85.42 / 6.0 \\
		\midrule
		\multirow{8}{*}{\rotatebox{90}{\textbf{Natural Images}}} & \textbf{Caltech-256} &&& $n_c=4$ & \parbox{2cm}{\centering$n_c=4$}\\
		& ResNet-50	            & 87.30 & 59.00 / 5.0 & 87.97 & 61.82 / 35.94\\
		& NFNet	                & 88.50 & 60.17 / 5.0 & 91.39 & 75.28 / 35.13\\
		& ConvNeXt-B	            & 91.57 & 78.32 / 5.0 & 93.74 & 79.86 / 14.85\\
		\cmidrule{2-6}
		& \textbf{RAF-Database} &&& $n_c=4$ & \parbox{2cm}{\centering$n_c=4$}\\
		& ResNet-50               & 81.91	& 67.42 / 5.0 & 80.82 & 64.43 / 13.23\\
		& NFNet	            & 83.96	& 69.68 / 5.0	& 82.82 &	69.75 / 13.33\\
		& ConvNeXt-B	            & 86.63	& 69.32 / 4.79 & 88.23 & 71.97 / 13.23\\
		\midrule
		\multirow{4}{*}{\rotatebox{90}{\textbf{Medical}}} & \textbf{ISIC} &&& $n_c=4$ & \parbox{2cm}{\centering$n_c=4$}\\
		& ResNet-50	            & 86.08 & 71.09 / 4.66 & 82.15 & 70.89 / 8.09 \\
		& NFNet	                & 90.16 & 77.23 / 14.61 & 86.90 & 71.39 / 18.28 \\
		& ConvNeXt-B	            & 87.20 & 69.63 / 14.02 & 81.87 & 68.78 / 30.69 \\
		\midrule
		\multirow{4}{*}{\rotatebox{90}{\textbf{Synthetic}}} & \textbf{SCDB} &&& $n_c=4$ & \parbox{2cm}{\centering$n_c=4$}\\
		& ResNet-50               & 90.20 & 86.20 / 4.60 & 92.19 & 85.13 / 60.00 \\
		& NFNet	                & 94.40 & 88.33 / 4.46 & 94.87 & 85.19 / 18.27 \\
		& ConvNeXt-B	            & 92.46 & 85.80 / 13.19 & 92.40 & 87.07 / 30.08 \\
		\bottomrule
	\end{tabular}
	}
\end{table}

Applying privacy-preserving training techniques for DL-based models can have a very diverse impact on their test performances.
The severity depends on multiple factors including model architecture, type of dataset, as well as the various hyperparameters for model and private training.
Table~\ref{tab:acc} shows the results on the respective test sets for all experiment configurations when trained with different private training techniques.
Results are sorted by domains and datasets to provide a better overview.

Even in privacy-preserving training settings, all models converged and demonstrated acceptable accuracies.
However, the best accuracies were usually achieved in \textit{Baseline} or \textit{FedAVG} settings.
Across all domains, it can be observed that \textit{DP} has a considerable impact on the models' test performances.
For some configurations, a higher $\epsilon$-value was required to achieve comparable results (e. g., \textit{NFNet} and \textit{ConvNeXt-B} for \textit{ISIC}).
However, no consistent pattern indicating higher robustness of one model architecture over another, against noise introduced by \textit{DP}, is obvious.
In contrast to \textit{DP}, \textit{FedAVG} always resulted in significantly lower performance losses.
The combination of \textit{FedAVG} and \textit{DP} almost exclusively resulted in a lower performance considering a comparable $\epsilon$-value.

The results from the time series domain for most datasets indicate that \textit{InceptionTime} is usually affected slightly less by private training, in direct comparison with \textit{ResNet-50}.
The only exception is the \textit{Anomaly} dataset, which experienced almost no performance loss with \textit{ResNet-50}.
One possible explanation for this is the advanced architecture of \textit{InceptionTime} including residual connections and inception modules. This enables the \textit{InceptionTime} to be more robust against noise and outliers.
All datasets in the time series domain, with the exception of \textit{Anomaly}, \textit{ECG5000}, and \textit{Wafer}, considerably suffered from the combination of \textit{DP} with \textit{FedAVG}.
For \textit{Character Trajectories} and \textit{FordA}, the $\epsilon$-value had to be increased to achieve adequate results.

The image domain indicates similar findings.
Since the models for the \textit{Tobacco3482} dataset were trained after being pretrained on the \textit{RVL-CDIP} dataset, the models were able to achieve higher performance even with \textit{DP} and \textit{FedAVG-DP}.
However, despite pretraining on \textit{Derm7pt}, the impact of \textit{DP} on \textit{ISIC}-trained models is still considerable.
For all datasets in the image domain, a moderately higher $\epsilon$-value was also experimented with, to improve the performance of the models for \textit{DP} and \textit{FedAVG-DP} cases.
However, it did not seem to provide any significant improvement in most of the cases.
It is worth noting that for larger datasets (i. e., \textit{RVL-CDIP}, \textit{Caltech-256}, \textit{RAF-Database} and \textit{ISIC}), \textit{DP} and \textit{FedAVG-DP} severely degraded the performance of the models whereas the performance for \textit{FedAVG} is still comparable to the \textit{Baseline} setting. 

\subsection{General Impact on Explainability (Qualitative)}

A visual inspection of individual explanations gives a first impression of the influence privacy-preserving techniques can have on the trained models.
These local impressions are then further validated on the dataset level through the qualitative analysis of summary statistics.

\subsubsection{Individual Analysis}
\begin{figure}[!t]
\centering
\subfigure[Anomaly Detection]{
\includegraphics[width=.98\linewidth]{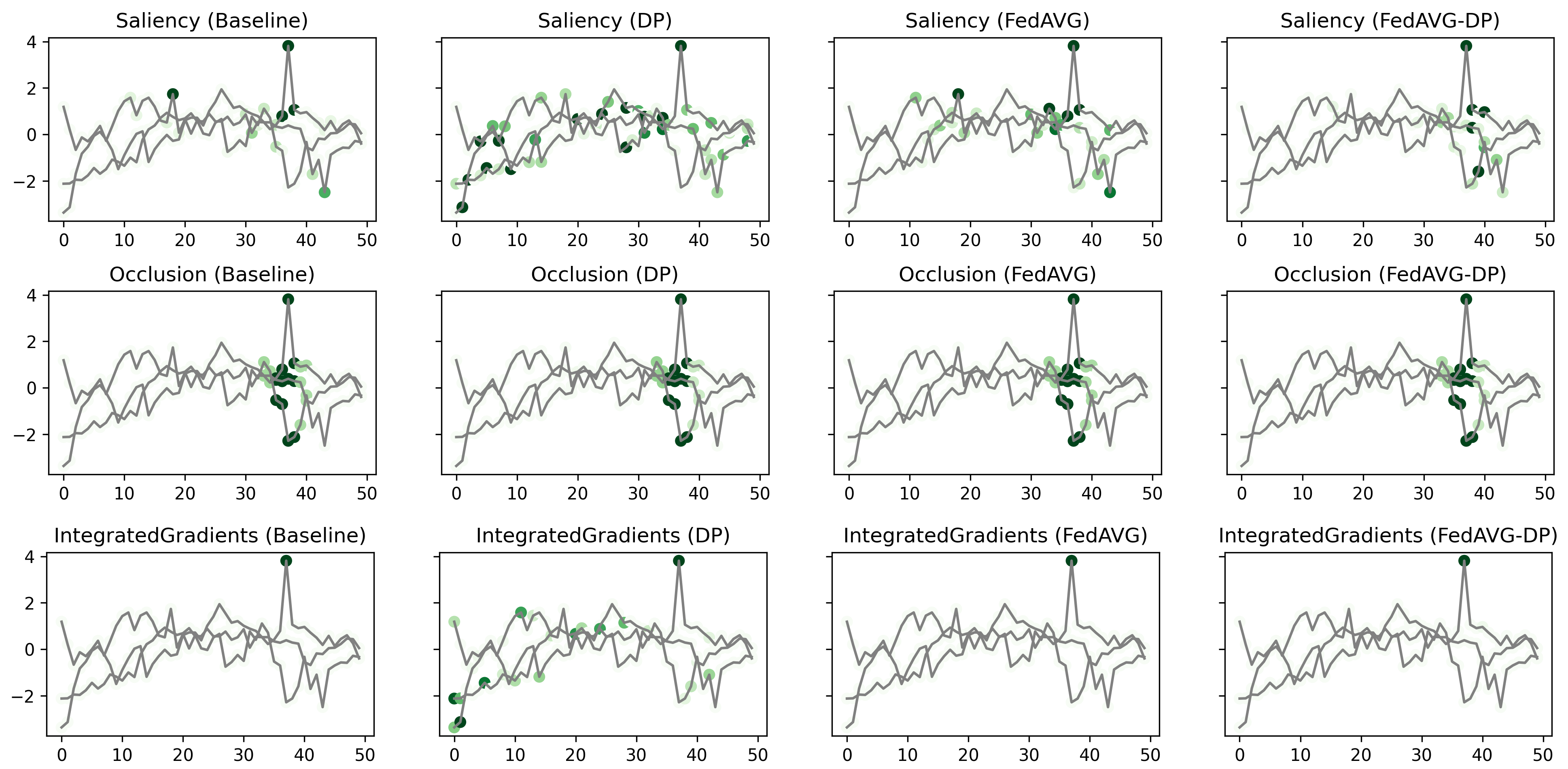}
\label{fig:qualitative_ts_anomaly}
}
\hfil
\subfigure[Character Trajectories]{
\includegraphics[width=.98\linewidth]{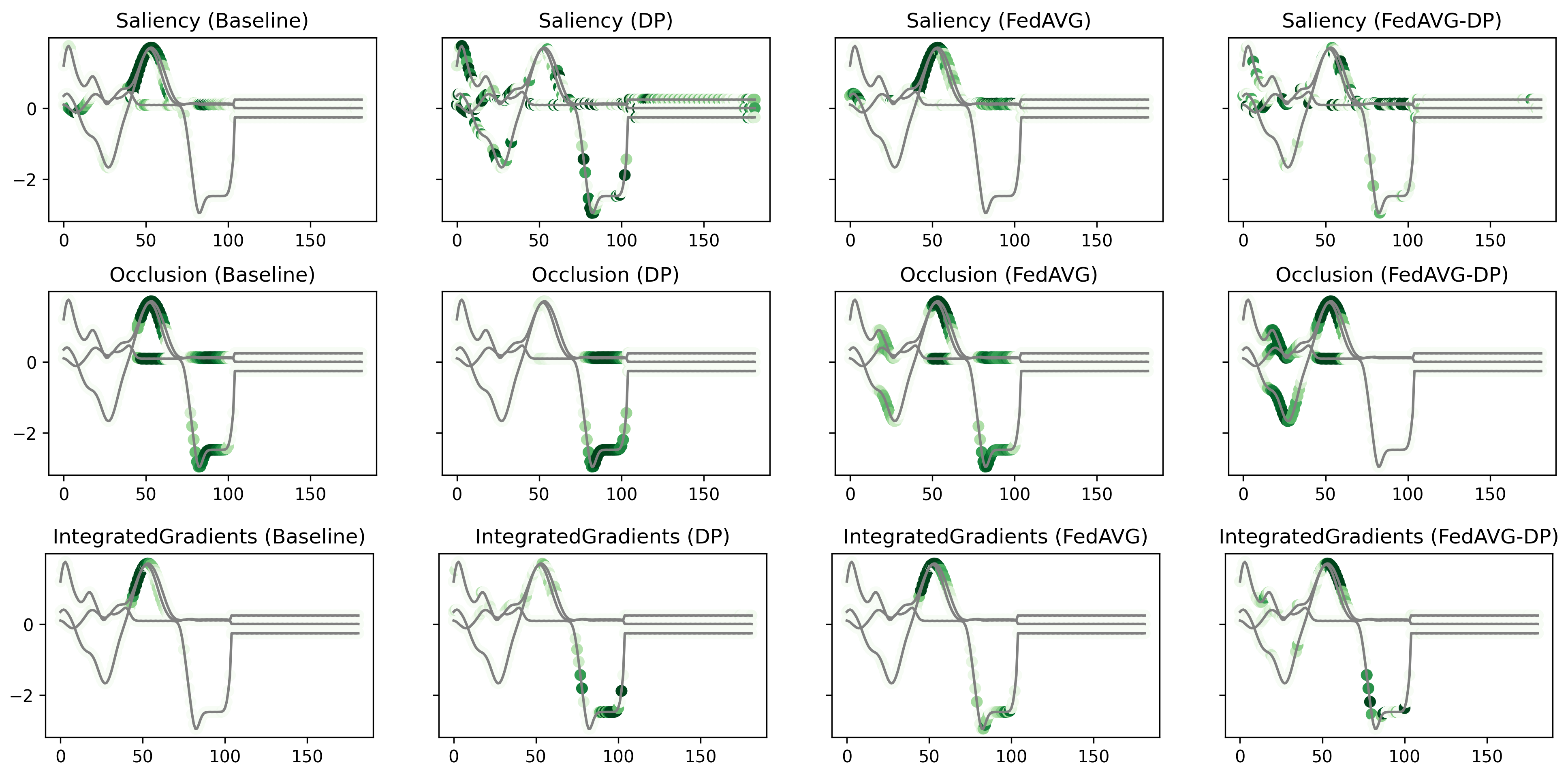}
\label{fig:qualitative_ts_character}
}
\caption{Shows the change in the attribution for \textit{ResNet-50} for three selected samples of the \textit{Anomaly Detection} and \textit{Character Trajectories} dataset, respectively. \textit{DP}-based training techniques tend to add additional noise and alter the explanation. \textit{FedAVG}, by contrast, is closer to the original attribution of the baseline.}
\label{fig:qualitative_ts}
\end{figure}

Figure~\ref{fig:qualitative_ts} shows explanations generated for the \textit{Anomaly Detection} and \textit{Character Trajectories} datasets in the time series domain.
For the \textit{Anomaly Detection} dataset, it can be observed that there is a general overlap between the explanations from different training settings, always highlighting the anomaly.
In some cases, \textit{DP} increases the amount of noise in the signal's relevance around the anomaly, yielding unclear and misleading explanations by highlighting distant points which do not correspond to the anomaly at all.
However, this is not the case when additionally adding \textit{FedAVG} in the \textit{FedAVG-DP} setting.
By contrast, \textit{DP}-trained models show remarkable deviations from the original \textit{Baseline} explanation when trained on the \textit{Character Trajectories} dataset.
This observations holds not only true for \textit{DP}, but also \textit{FedAVG-DP} settings.
\textit{FedAVG}, on the other hand, shows explanations close to the \textit{Baseline} setting, with only minor deviations.

\begin{figure*}[!t]
\centering
\includegraphics[width=0.8\linewidth]{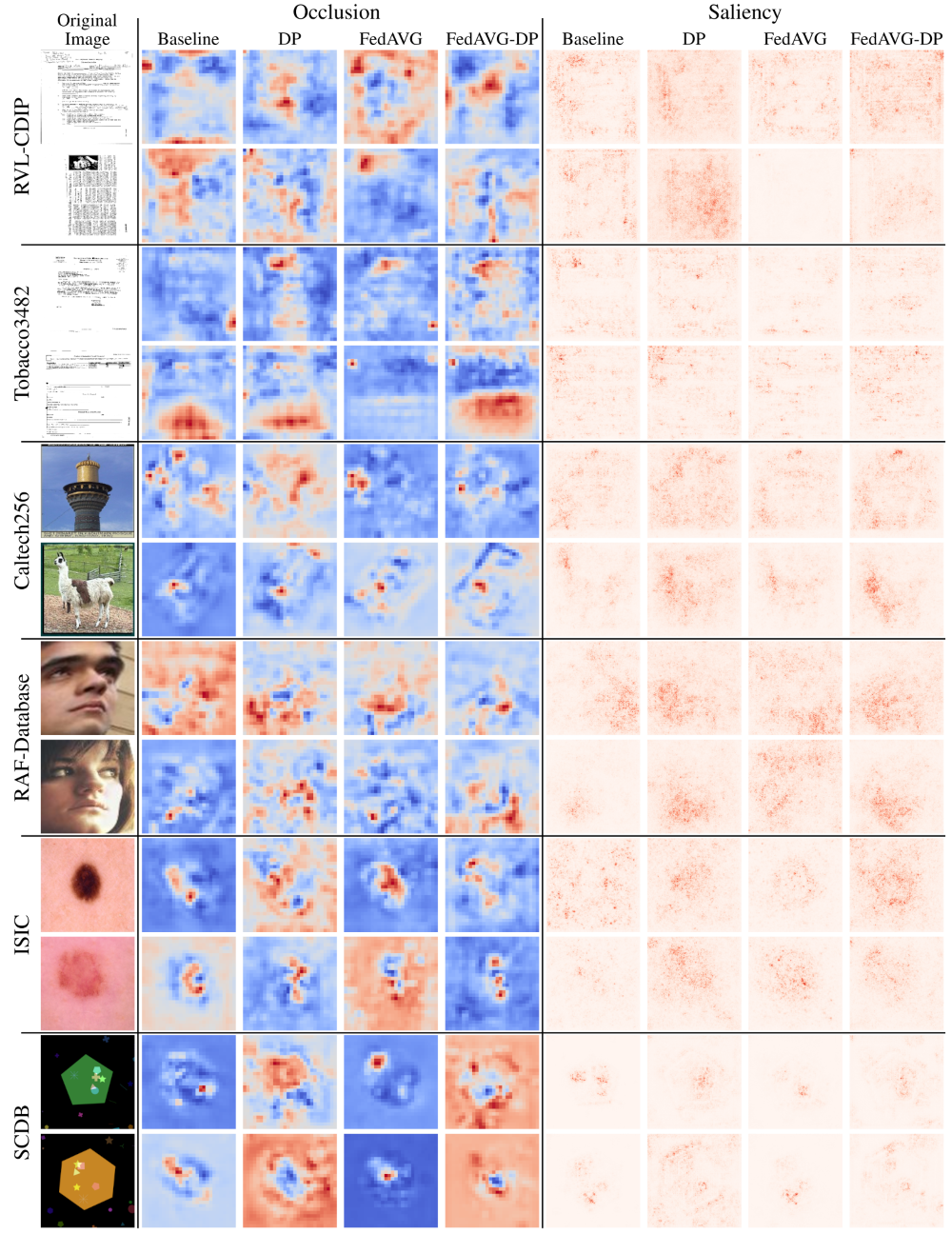}
\caption{Examples of perturbation- and gradient-based attribution maps computed on models trained in varying privacy configurations. Two random, correctly classified samples are provided per dataset. \textit{Occlusion} and \textit{Saliency} attribution maps are computed on \textit{ResNet-50} models. Red and blue regions indicate positive and negative relevance, respectively.}
\label{fig:qualitative_image}
\end{figure*}

Figure~\ref{fig:qualitative_image} shows samples from all image datasets along with the generated \textit{Occlusion} and \textit{Saliency} explanations from \textit{ResNet-50} models trained with and without privacy techniques.
It can be seen that different training settings yield heatmaps that visibly differ.
However, the areas of highest relevance roughly overlap for most samples.
Particularly for \textit{Saliency} attributions, it can be observed that \textit{DP} and \textit{FedAVG-DP} often add additional noise to the generated explanations.
In some cases, this can also be observed in \textit{Occlusion}.
This is particularly striking in the \textit{SCDB} samples in the last two rows, where both \textit{DP}-based methods highlight regions outside the decision-relevant area.
Another striking example is the second sample from \textit{RVL-CDIP}.
It can be seen that both \textit{FedAVG} attribution maps present smoother and more focused heatmaps pointing to a specific location on the image.
On closer inspection of the samples from document datasets, it was found that \textit{FedAVG} prominently focused on specific class-relevant cues such as dates, titles, figures, etc.
Moreover, comparing \textit{DP} with \textit{FedAVG-DP} attribution maps, it can be observed that the addition of federated training leads to less noise in the attribution for some samples.
When observing the \textit{SCDB} sample in the last row, it can be observed that for \textit{Occlusion}, \textit{Baseline} highlights both rectangle and star shapes, whereas \textit{FedAVG} only focuses on the rectangle shapes.
In \textit{SCDB}, rectangles are exclusive markers for class two.
However, both star and star markers are also part of the decision-relevant shape combination.
\textit{FedAVG} appears to have focused only on the single relevant marker, whereas in the \textit{Baseline} setting, multiple relevant markers were highlighted.
This is also evident from the \textit{SCDB} sample in the second last row, where \textit{FedAVG} successfully focused on the ellipse shape, which is exclusive for class one. 
It has to be noted that the \textit{Occlusion} attribution map shows two relevant regions, where the most relevant region highlights the edge of the big, green pentagon.
This could be attributed to a limitation of \textit{Occlusion}, corresponding to distraction due to the generation of out-of-distribution samples during perturbation. 
Interestingly, \textit{Occlusion} attribution maps for \textit{Baseline} and \textit{FedAVG} show different decision-relevant cues for the last row's sample, as compared to the corresponding \textit{Saliency} maps.
The former highlight rectangle and star shapes, whereas the latter most prominently highlight the star marker on the lower part of the image.
As already mentioned, star markers are also decision-relevant for that sample.
However, they are no exclusive markers and could also indicate shape combinations appearing in class one.
Nevertheless, it has to be mentioned that for some individual samples, these observations do not apply.
For the second sample of \textit{ISIC} and the first sample of \textit{RAF-Database}, for instance, \textit{Baseline} and \textit{FedAVG} did not yield clearer heatmaps as compared to the \textit{DP}-based approaches.
To capture overall trends and characteristics in attribution maps of different configurations, further analysis on dataset-level statistics of the generated attribution maps was performed.

\begin{figure}[!t]
\centering
\subfigure[InceptionTime]{
\includegraphics[width=.46\linewidth]{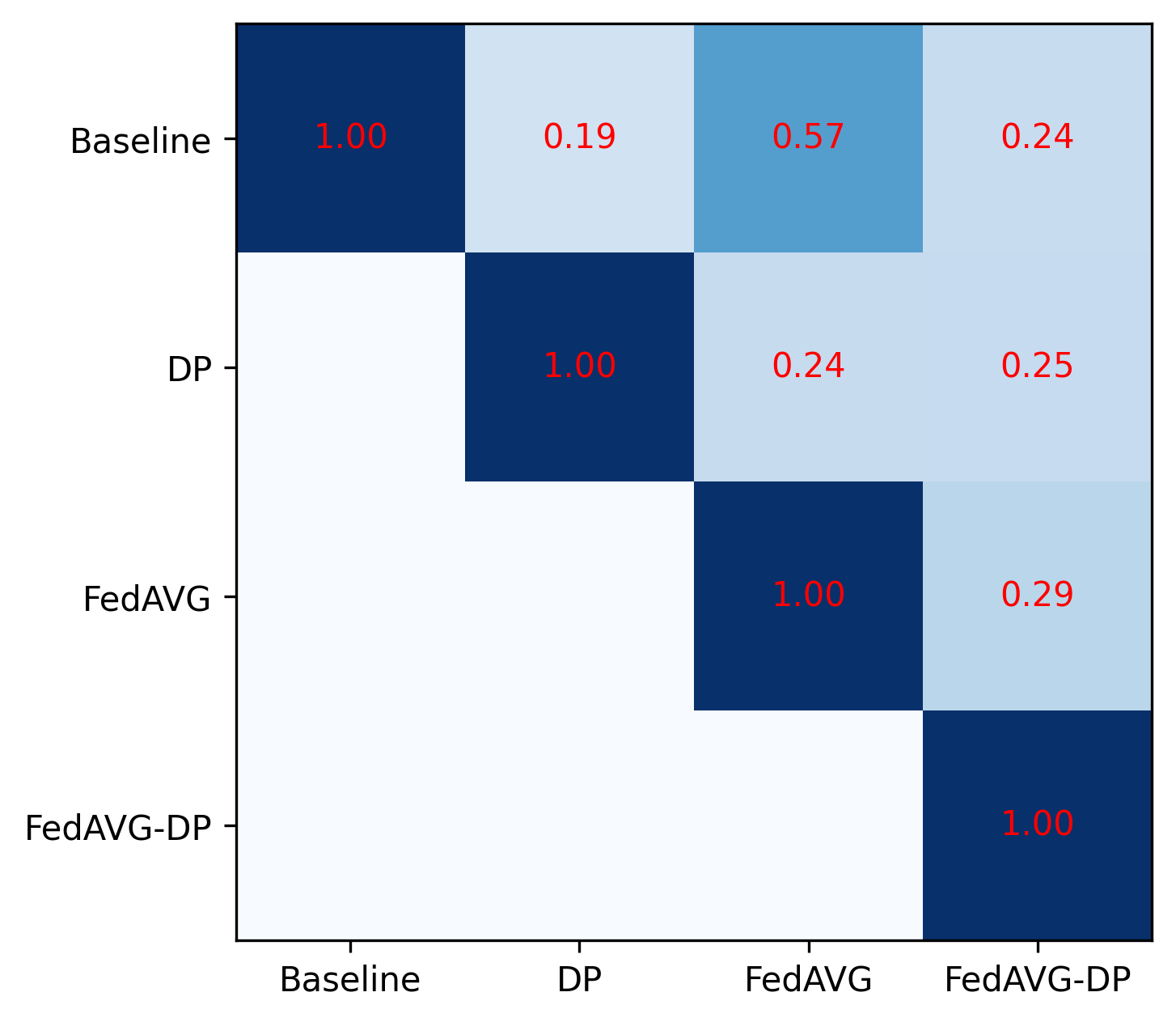}
\label{fig:corr_ts_inception}
}
\hfil
\subfigure[ResNet]{
\includegraphics[width=.46\linewidth]{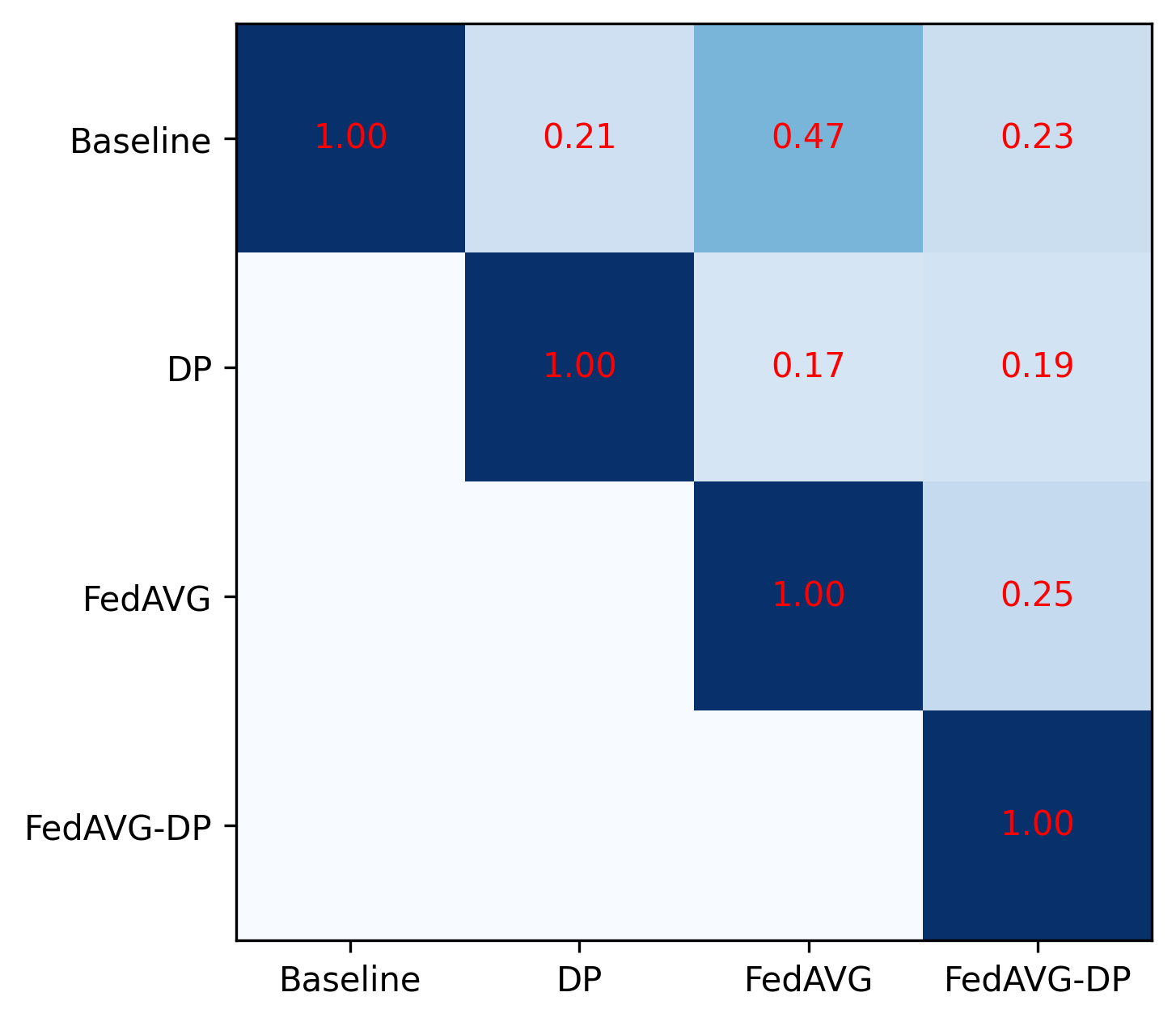}
\label{fig:corr_ts_resnet}
}
\caption{Shows the average Pearson correlation of the attribution maps compared between the different privacy approaches for the Anomaly Detection dataset. \textit{FedAVG} shows a higher similarity to the \textit{Baseline} setting, as compared to the \textit{DP}-based approaches.}
\label{fig:corr_ts}
\end{figure}

\subsubsection{Dataset-wide analysis}
Figure~\ref{fig:corr_ts} shows the Pearson correlation of the explanations generated by different training settings for the \textit{Anomaly Detection} dataset. 
Therefore, the correlation across the different training approaches was computed using all available attribution maps. 
Precisely speaking, the attribution between the corresponding attribution maps was calculated and the average over the number of samples was taken.
The final correlation shows the score averaged over the attribution methods and the samples. 
For both architectures, it is evident that the privacy methods significantly change the produced attribution maps. 
However, \textit{FedAVG} yields significantly higher correlation to the \textit{Baseline} setting as compared to the \textit{DP}-based approaches.
Moreover, it is surprising that the correlation between \textit{DP} and \textit{FedAVG-DP} is rather low.
The remaining correlation matrices are excluded as they showed similar results to the presented and do not provide any additional information.

\begin{figure}[!t]
\centering
\includegraphics[width=\linewidth]{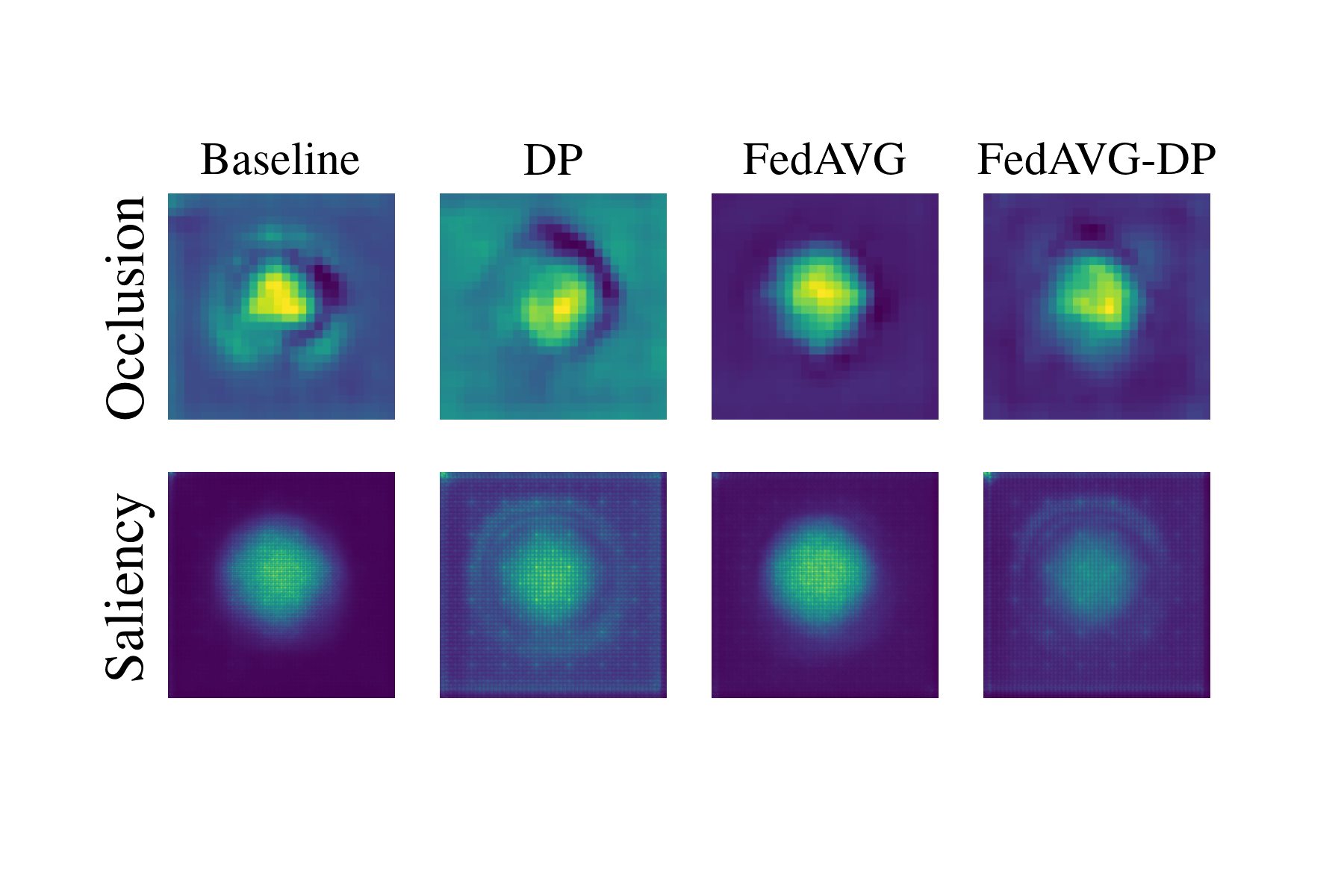}
\caption{Average attribution heatmaps for samples from \textit{SCDB} computed for each of the non-private / private approaches.}
\label{fig:avg_heatmaps_medical}
\end{figure}

\textit{SCDB} is a synthetic dataset that, by design, carries all relevant information in the center of the image.
Figure~\ref{fig:avg_heatmaps_medical} shows a visual comparison of the overall impact of privacy on the distribution of attribution in the explanations.
For each setting, the attribution maps are averaged across all test samples to highlight the frequency with which regions were attributed as relevant.
It is evident that involving \textit{DP} during training drastically increased the diffusion of attribution values, also including image areas that are not related to the actual classification task.
Combining \textit{DP} with \textit{FedAVG}, on the other hand, led to a moderation of the noise added by \textit{DP}.
Interestingly, the results show that \textit{FedAVG} alone usually results in the cleanest and most focused heatmaps, even improving the baseline.
As the remaining datasets possess less spatial standardization, it is not trivial to interpret their results in the same way.

This qualitative analysis already drew an interesting initial picture, proving that to some degree, any privacy-preserving training technique has an impact on the generated explanations.
Furthermore, the results suggest that \textit{DP}-trained models generate explanations that tend to be noisier and cover potentially unimportant regions, harboring the danger of misleading the explainer.
However, it is unclear whether noise added by \textit{DP} only concerns the explanations, or whether this reflects the model decision (Fidelity).
First results also indicate that the \textit{FedAVG} approach can even improve explanations, leading to more focused, and meaningful explanations in some instances.

\subsection{General Impact on Explainability (Quantitative)}
\label{sec:quantitative_impact}
The quantitative analysis serves as a means to further verify the findings from the previous section and allows the investigation of whether privacy-preserving techniques impact only the explanations, or also the underlying model behavior.

\subsubsection{Continuity}

\begin{figure}[!t]
\centering
\subfigure[Time series datasets]{
\includegraphics[width=.98\linewidth]{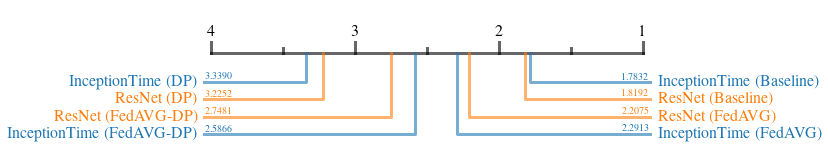}
\label{fig:cd_con_time}
}
\subfigure[Document image datasets]{
\includegraphics[width=.98\linewidth]{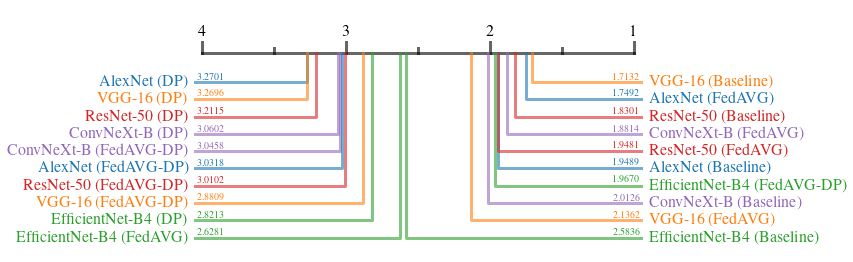}
\label{fig:cd_con_doc}
}
\subfigure[Natural image datasets]{
\includegraphics[width=.98\linewidth]{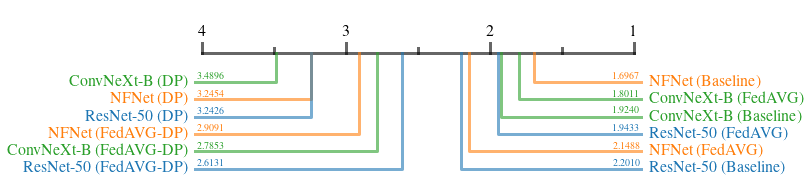}
\label{fig:cd_con_nat}
}
\subfigure[Medical image datasets]{
\includegraphics[width=.98\linewidth]{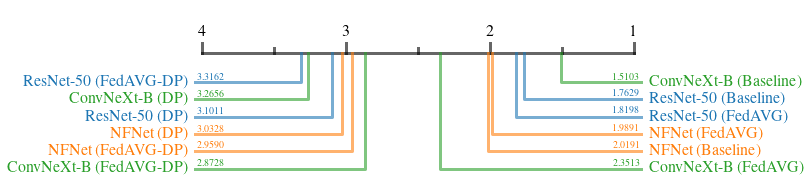}
\label{fig:cd_con_med}
}
\subfigure[Synthetic image datasets]{
\includegraphics[width=.98\linewidth]{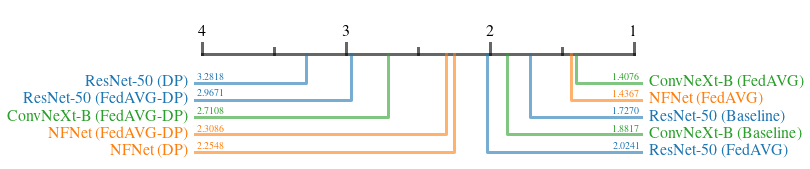}
\label{fig:cd_con_syn}
}
\caption{Critical difference diagrams for the Continuity of models trained on datasets from different domains. Privacy results in less continuity and therefore noisier explanations.}
\label{fig:cd_con}
\end{figure}

Measuring the continuity of an explanation helps to understand how difficult the interpretation of an explanation might be for an explainer. 
Humans usually struggle when confronted with high-dimensional, diffuse data.

The continuity for time series data is defined as the sum of the absolute changes between each pair of subsequent points within the attribution map. 
For image attributions, continuity is computed as the sum of absolute gradients in both spatial directions of the attribution map.
A smoother map results in a lower continuity.
Figure~\ref{fig:cd_con} shows the CD diagrams for the \textit{Continuity} across all domains.
For each domain, the ranked results are averaged over all datasets and attribution methods.

The results for all domains clearly show that \textit{Baseline} and \textit{FedAVG} settings yield better continuity scores as compared to the \textit{DP}-based approaches.
This confirms that \textit{DP}-based private models generate significantly more discontinuous attribution maps compared to \textit{Non-DP} training techniques.
The only outlier to this observation is \textit{EfficientNet-B4} trained on document images in Figure~\ref{fig:cd_con_doc}, where surprisingly \textit{FedAVG-DP} achieved the highest rank.
Comparing only \textit{Baseline} and \textit{FedAVG} models, it can not be clearly stated whether one is better than the other, as this seems to be highly dependent on the exact model architecture and domain combination.
Moreover, \textit{FedAVG-DP} achieved better ranks as compared to \textit{DP} in most configurations across all domains.
\textit{DP} and \textit{Non-DP} approaches even show a clear visual separation in the CD diagram in most cases (i. e., document, natural, and medical).
For document image datasets, there are only minor differences within the ranks of \textit{DP} and \textit{Non-DP} regions, making it very hard to draw clear conclusions.

\subsubsection{Area over the Perturbation Curve}

\begin{figure}[!t]
\centering
\subfigure[Time series datasets]{
\includegraphics[width=.98\linewidth]{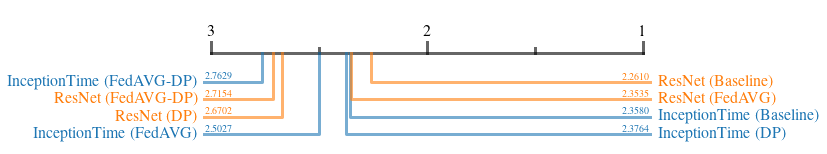}
\label{fig:cd_aopc_time}
}
\subfigure[Document image datasets]{
\includegraphics[width=.98\linewidth]{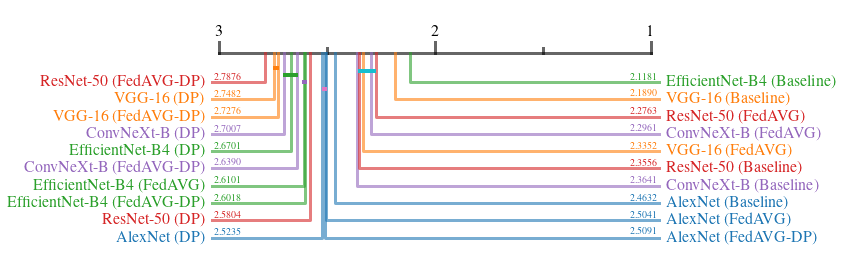}
\label{fig:cd_aopc_doc}
}
\subfigure[Natural image datasets]{
\includegraphics[width=.98\linewidth]{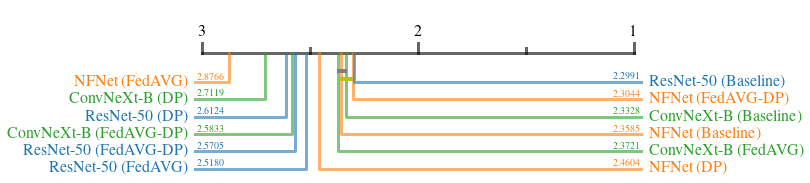}
\label{fig:cd_aopc_nat}
}
\subfigure[Medical image datasets]{
\includegraphics[width=.98\linewidth]{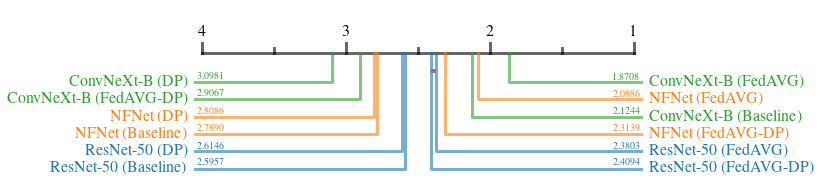}
\label{fig:cd_aopc_med}
}
\subfigure[Synthetic image datasets]{
\includegraphics[width=.98\linewidth]{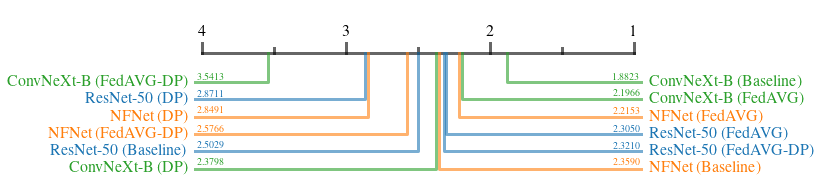}
\label{fig:cd_aopc_syn}
}
\caption{Critical difference diagrams for the AOPC of models trained on datasets from different domains.}
\label{fig:cd_aopc}
\end{figure}

The \textit{AOPC} measures how removing features deemed relevant by the explanation affects local model predictions.
This provides important insights into the Fidelity of the explanations.
Intuitively, removing features with lower importance should affect the prediction less, whereas the deletion or perturbation of important features should result in significant prediction changes. 
In this experiment, features were removed sequentially starting with the most important, as per the attribution map.

Figure~\ref{fig:cd_aopc} shows all critical difference diagrams for the \textit{AOPC} measure.
Over all domains, the most prevalent pattern is that of \textit{Non-DP}-based settings occupying the higher ranks.
In the time series domain, \textit{ResNet-50} shows the clear superiority of \textit{Baseline} and \textit{FedAVG} compared to \textit{DP} and \textit{FedAVG-DP}.
However, the results from the image domain indicate that a clear superiority of \textit{FedAVG} over \textit{Baseline} can not be reported.
In contrast to other domains, \textit{FedAVG}-based approaches on average achieved higher ranks on medical and synthetic images.
For \textit{InceptionTime}, \textit{DP} surprisingly achieved almost similar performance as compared to \textit{Baseline}.
Apart from this outlier, \textit{DP} almost exclusively ranked last in direct comparison with all other training settings.
The presented results suggest that adding \textit{Differential Privacy} during training decreases the explanation's fidelity.

\subsubsection{Infidelity}

\begin{figure}[!t]
\centering
\subfigure[Time series datasets]{
\includegraphics[width=.98\linewidth]{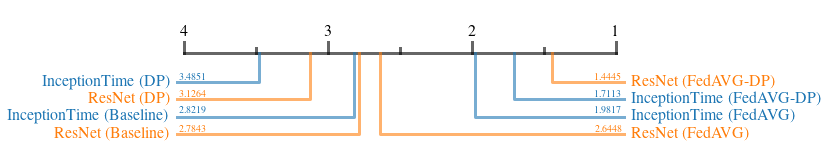}
\label{fig:cd_inf_time}
}
\subfigure[Document image datasets]{
\includegraphics[width=.98\linewidth]{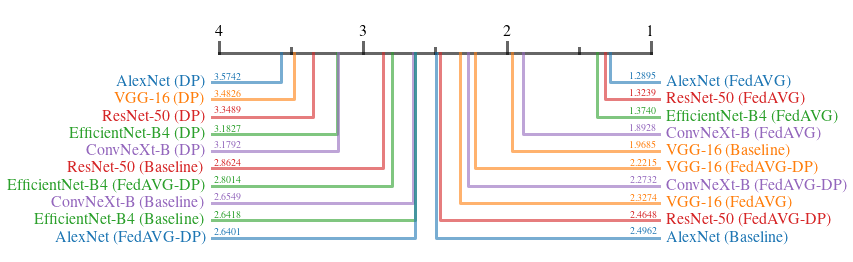}
\label{fig:cd_inf_doc}
}
\subfigure[Natural image datasets]{
\includegraphics[width=.98\linewidth]{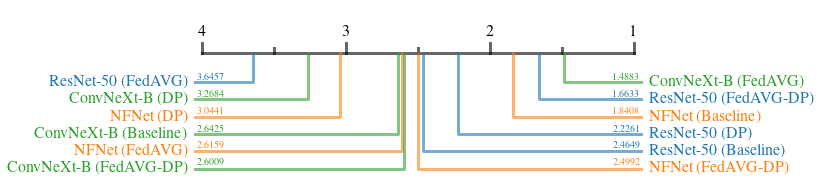}
\label{fig:cd_inf_nat}
}
\subfigure[Medical image datasets]{
\includegraphics[width=.98\linewidth]{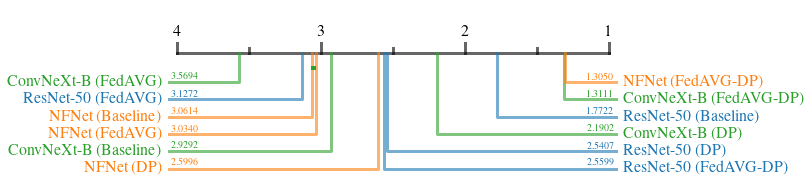}
\label{fig:cd_inf_med}
}
\subfigure[Synthetic image datasets]{
\includegraphics[width=.98\linewidth]{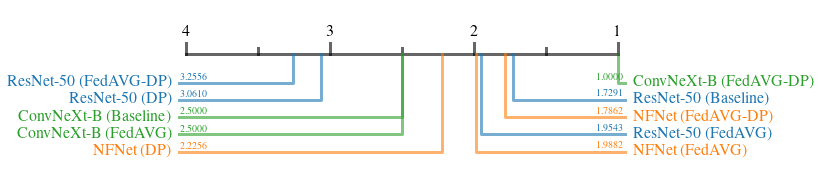}
\label{fig:cd_inf_syn}
}
\caption{Critical difference diagrams for the Infidelity of models trained on datasets from different domains.}
\label{fig:cd_inf}
\end{figure}

The \textit{Infidelity} measure provides information about an explanation's fidelity by evaluating a model's adversarial robustness in regions of varying explanation relevance. 
Perturbations are both applied to the attribution map and the input image while comparing the predictions of the unperturbed and noisy input.
It is expected that the perturbation of a more important feature leads to a larger change in prediction. 

Figure~\ref{fig:cd_inf} shows the \textit{Infidelity} ranks for all configurations.
In the time series domain, approaches involving \textit{FedAVG} achieved the highest scores.
Interestingly, the addition of \textit{DP} to \textit{FedAVG} settings resulted in higher scores, whereas the sole use of \textit{DP} during training led to the worst outcomes.
For all image domains, again, results depend on the architecture and the domain.
However, there is an overall tendency similar to the results of the time series domain with \textit{DP} being the worst.
Furthermore, \textit{FedAVG} approaches being the best performing approach for time series datasets.
The \textit{Baseline} setting ranked highest in several configurations, such as \textit{ResNet-50} in medical and synthetic images, and \textit{NFNet} in natural  images.
This indicates that \textit{FedAVG} increases adversarial robustness and Fidelity while \textit{DP} alone leads to lower Fidelity.

\subsubsection{Sensitivity}
\begin{figure}[!t]
\centering
\subfigure[Time series datasets]{
\includegraphics[width=.98\linewidth]{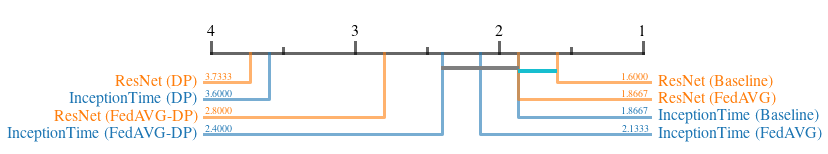}
\label{fig:cd_sens_time}
}
\subfigure[Document image datasets]{
\includegraphics[width=.98\linewidth]{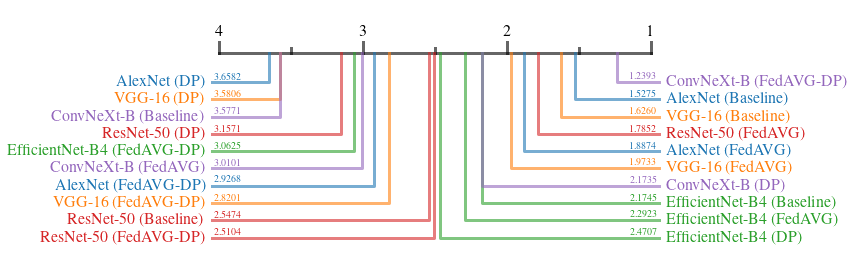}
\label{fig:cd_sens_doc}
}
\subfigure[Natural image datasets]{
\includegraphics[width=.98\linewidth]{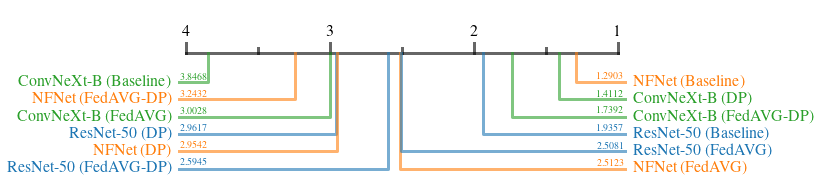}
\label{fig:cd_sens_nat}
}
\subfigure[Medical image datasets]{
\includegraphics[width=.98\linewidth]{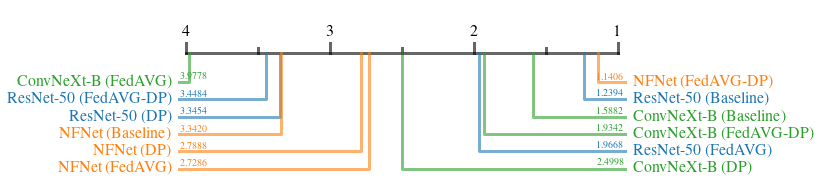}
\label{fig:cd_sens_med}
}
\subfigure[Synthetic image datasets]{
\includegraphics[width=.98\linewidth]{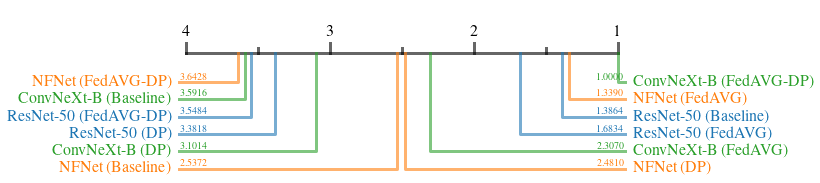}
\label{fig:cd_sens_syn}
}
\caption{Critical difference diagrams for the Sensitivity of models trained on datasets from different domains.}
\label{fig:cd_sens}
\end{figure}

In contrast to the \textit{Infidelity}, \textit{Sensitivity} quantifies the fidelity by perturbing the input directly.
The change in the generated explanation is measured before and after the input is insignificantly perturbed.
Small changes in the input should not result in large changes in the attribution map.

Figure~\ref{fig:cd_sens} shows the \textit{Sensitivity} ranks for all configurations.
For the time series domain, Figure~\ref{fig:cd_sens_time} shows a very clear ranking with \textit{Baseline} and \textit{FedAVG} being superior to \textit{FedAVG-DP}, followed by \textit{DP}.
However, no clear statistical distinction can be made between \textit{Baseline}, \textit{FedAVG}, and \textit{FedAVG-DP} for \textit{InceptionTime}, and for \textit{Baseline} and \textit{FedAVG} for \textit{ResNet-50}.
The superiority of \textit{Non-DP}-based over \textit{DP}-based approaches is further confirmed by seven more configurations within the different image domains, including \textit{ResNet-50} in natural, medical, and synthetic images.
Interestingly, both \textit{DP}-based methods ranked highest in combination with \textit{ConvNeXt} in all image-domains except for medical imaging.

\subsubsection{Ground Truth Concordance}
Measuring the concordance of attribution maps and ground truth explanations is the best way to ensure the truthfulness of explanations, but comes with some limitations.
\textit{Ground Truth Concordance} can only be computed on synthetically constructed datasets without ambiguous decision paths.
Therefore, the \textit{SCDB} dataset was utilized which provides segmentation maps for the different visible shapes to construct ground truth explanation maps containing only decision-relevant shapes for each image.
All attribution maps are blurred before computing the concordance, to moderate the drawback of gradient-based methods, which generate noisier explanations by design.

\begin{figure}[!t]
\centering
\includegraphics[width=\linewidth]{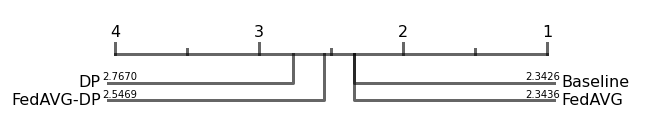}
\caption{Critical difference diagram for \textit{Ground Truth Concordance} on SCDB data for different privacy-preserving settings. \textit{FedAVG} and \textit{Baseline} settings clearly outperform \textit{DP}-based techniques.}
\label{fig:cd_gtc}
\end{figure}

Figure~\ref{fig:cd_gtc} shows the critical difference diagram for the \textit{Ground Truth Concordance} computed over all attribution methods for the \textit{SCDB} dataset.
The ranking clearly indicates the superiority of \textit{Baseline} and \textit{FedAVG} settings over \textit{DP}-based training techniques.
Moreover, it can be observed that the addition of \textit{FedAVG} to the \textit{DP}-trained setting alleviates the divergence from the ground truth explanations.
Overall it can be noted that \textit{DP}-based methods indeed reduce the fidelity of models, while there is promising evidence that a \textit{DP}-based training in a federated constellation can mitigate its effects to a certain degree.

\subsection{Impact of Noise on Different Settings}
\label{sec:impact_of_noise}
The results so far suggested that the introduction of \textit{DP} during the training process has a considerable impact on the generated explanations.
Moreover, it was found that using the combination of \textit{FedAVG} and \textit{DP} can sometimes mitigate the negative effects of the added noise during the training process.
This section covers an investigation whether the degree to which the quality of explanations is affected, differs for different attribution methods and datasets.
Therefore, the relative increase in continuity score was measured when comparing the \textit{Baseline} with the \textit{DP} training setting.
A higher relative increase indicates a bigger impact, resulting in a lower rank.

\subsubsection{Attribution Methods}
\begin{figure}[!t]
\centering
\subfigure[Time series datasets]{
\includegraphics[width=.98\linewidth]{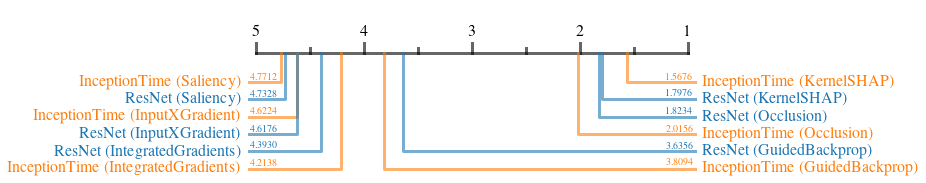}
\label{fig:cd_noa_time}
}
\subfigure[Image datasets]{
\includegraphics[width=.98\linewidth]{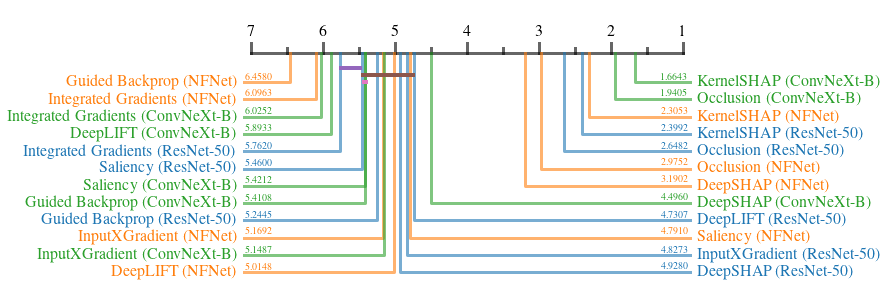}
\label{fig:cd_noa_image}
}
\caption{Critical difference diagrams showing the impact of adding \textit{Differential Privacy} during training, on the quality of explanations generated by different attribution methods.}
\label{fig:cd_noa}
\end{figure}

Figure~\ref{fig:cd_noa} shows the ranks of different attribution methods when applied to different architectures before and after adding \textit{DP} to the training, for the time series and image datasets.
For both modalities, a prominent separation of two distinct groups can be noticed.
In time series datasets, both \textit{KernelSHAP} and \textit{Occlusion} are affected significantly less by differential privacy as compared to the remaining, gradient-based methods.
Similarly, \textit{KernelSHAP} and \textit{Occlusion} clearly outperformed most other methods.
\textit{DeepSHAP} is the only exception, which even achieved a similar score to \textit{Occlusion} when applied to \textit{NFNet}.

\subsubsection{Datasets}
\begin{figure}[!t]
\centering
\subfigure[Time series datasets]{
\includegraphics[width=.98\linewidth]{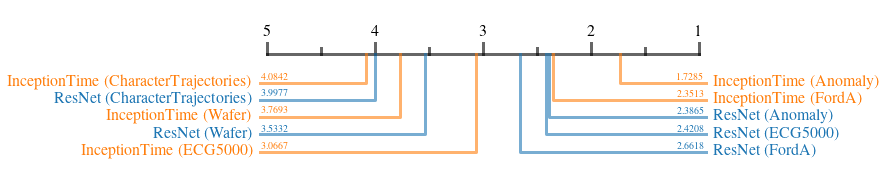}
\label{fig:cd_nod_time}
}
\subfigure[Image datasets]{
\includegraphics[width=.98\linewidth]{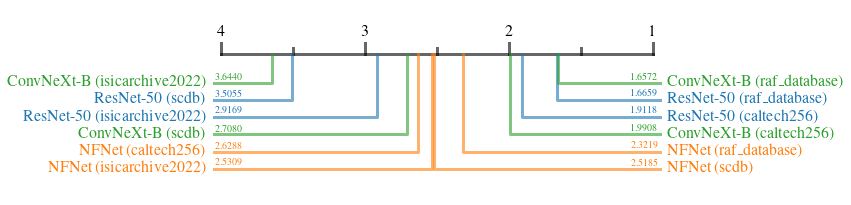}
\label{fig:cd_nod_image}
}
\caption{Critical difference diagrams showing the impact of adding \textit{Differential Privacy} during training, on the quality of explanations when applied to different datasets.}
\label{fig:cd_nod}
\end{figure}

Figure~\ref{fig:cd_nod} shows the impact of \textit{DP} on the quality of explanations for different datasets.
For the time series domain, it can be seen that noise has the least impact on the \textit{Anomaly} dataset, as the decision-relevant anomaly is not affected much by the added noise.
On the other hand, \textit{Character Trajectories} dataset is highly affected by noise.
This can be explained by the fact that the dataset consists of raw sensor values that describe drawn letters.
Slight noise distributed over the time series can have a devastating influence on the meaning of a given sample, as the error adds up over time.
In the image-domain, \textit{RAF-Database} and \textit{Caltech-256} are influenced less by noise, whereas \textit{ISIC}, on average, shows a higher susceptibility.
This is understandable, as \textit{ISIC} heavily relies on fine-grained patterns and complex features which might be more susceptible to added noise as compared to coarse-grained features used for emotion recognition and object detection. 
Surprisingly, the results suggest a rather high impact on \textit{SCDB} as well.
At first, this might seem unexpected due to the relevance of clean and uniform shapes for classification.
However, considering the low resolution of input images, these shapes might be particularly fragile under the influence of noise, as small perturbations can easily shift the resemblance of one shape to another.

\section{Discussion}
\label{sec:discussion}

In the last few years, explainability and data privacy are drastically gaining importance in the field of Deep Learning.
It is therefore all the more important to take a closer look at their interaction.
The presented results revealed a significant impact of privacy-preserving training techniques on generated explanations.
However, the influence on XAI strongly depends on the privacy technique used, as well as further factors.

First of all, it has been shown that not every PPML method has the same impact on model performance.
\textit{DP}-based models were shown to almost always deteriorating test accuracy.
Moreover, experience showed that they drastically complicate model convergence and hyperparameter search.
\textit{FedAVG}, on the other hand, yielded accuracies similar to the \textit{Baseline} setting, sometimes even improving the results.
It has to be mentioned, though, that both \textit{DP} and \textit{FedAVG} follow different goals in the domain of privacy.
Whereas \textit{DP} aims at preventing models to capture individual sample information, which could be used for reconstruction, \textit{FedAVG} mainly aims at minimising the exposure of sensitive information by keeping the training data local.
Although \textit{FedAVG} also generates an aggregated model which might have less vulnerability to reconstruction attacks due to averaging effects, it still needs to transfer information about the local models to the orchestration server.
Therefore, the combination of \textit{FedAVG} and \textit{DP} provide the highest privacy, yielding in many cases similar performance compared to only \textit{DP}.

The qualitative and quantitative analysis revealed various interesting findings regarding the impact of different privacy-preserving techniques on explanations.
\textit{Differential Privacy}, for example, stood out in almost all configurations for its property to add noise to the attribution maps.
This has been reported in many individual samples and could be confirmed by dataset level analysis, as well as quantitative analysis, where \textit{DP}-based methods stood out for increased \textit{Continuity} values.
One possible reason for this phenomenon is the addition of noise during the training process with \textit{DP}.
The introduction of noise in the parameter update most likely leads to contortions in the parameter space, which are never completely compensated, and translate into the prediction process.
This effect might be counteracted by slightly tweaking the optimization, such as fine-tuning public datasets, or by increasing the batch sizes during training.

The degree to which noise is added has been investigated in Section~\ref{sec:impact_of_noise}.
The results suggest, that perturbation-based methods are a lot less prone to changing their explanation's \textit{Continuity} under influence of noise.
Inspecting the individual examples, as well as the average heatmaps in Figure~\ref{fig:avg_heatmaps_medical}, this finding can again be verified.
The difference between \textit{Occlusion} and \textit{Saliency} is particularly notable in the contrast between highly relevant areas and low relevant areas.
Whereas \textit{Saliency} produces monotonous heatmaps, peaks and areas of interest are much more prominently highlighted in the average \textit{Occlusion} maps.
The main reason for perturbation-based methods being less affected by noise in terms of \textit{Continuity} is their higher resolution which neglects fine nuances in relevance, and the fact that randomly introduced noise is prone to cancel out within a patch.
However, \textit{Continuity} is only a mathematical approximation of an explanation's interpretability.
Figures~\ref{fig:qualitative_ts_character} and \ref{fig:qualitative_image} illustrate that \textit{Occlusion}-based explanations are often significantly changed when introducing \textit{DP} during training.
Furthermore, the high interpretability of heatmaps is worthless, if their fidelity is not ensured.
As reported in Section~\ref{sec:quantitative_impact}, \textit{DP} exclusively led to the deterioration of metrics indicating an explanation method's fidelity.
Therefore, even when applying \textit{Occlusion}, it needs to be clarified how truthful the generated explanations remain to be.

In contrast to \textit{DP}, \textit{Federated Learning} often resulted in smoother attribution.
Interestingly, combining \textit{FedAVG} with \textit{DP} often times even led to more continuous attribution maps compared to the \textit{Baseline} setting, reducing the negative effects introduced by \textit{DP} alone.
However, \textit{FedAVG-DP} has also been reported to decrease the fidelity of explanations in many cases.
Therefore, whenever XAI is required and \textit{Differential Privacy} is applied, it might be worth considering a combination of \textit{DP} and \textit{Federated Learning}.
This will also be possible in cases where \textit{Federated Learning} is not required, as the federated setting can easily be simulated by dividing the dataset into chunks.
Although some outlier experiments report a better \textit{Continuity} score for \textit{Baseline} settings, the fact that \textit{FedAVG} leads to better \textit{Continuity} scores has a strong theoretical basis.
Averaging models during training inevitably prevents the final model from overemphasizing granular features or noise.

The present study also showed that the influence of PPML on XAI is not really dependent on the application domain, but rather on the choice and feature scales of the dataset at hand.
The noise introduced by \textit{DP} has, above all, a detrimental impact on classification tasks that rely on fine-grained and nuanced features or patterns.
For simpler anomaly detection tasks or tasks focusing on the detection of overall, coherent structures seem to be less affected by privacy-preserving training techniques.

Besides the different influences PPML has on XAI, there is another fact that needs to be considered when combining both techniques.
No matter how private a system has been made, exposing an explanation is in itself always a potential point of attack for a system, revealing sensitive information about the decision-making process.
This is, for instance, particularly evident with \textit{Saliency}, which provides the raw gradients of a single input instance. 
For truly critical applications one should ask the question of who, in the end, should be authorized to request explanations, and under which circumstances.
Moreover, it might even be required to further obfuscate the exact generation process for explanations, or rely exclusively on global explanations for applications with extremely high privacy requirements.

This study revealed several general trends which will affect explanations on a global scale when applying private training strategies to DL-based models.
However, one major limitation of such studies is the examined basis of comparison.
When comparing explanations of separate model instances, there is always the risk of obtaining different local minima, i.e., different classification strategies.
Previous research~\cite{ilyas2019adversarial} suggests that one dataset can have multiple, redundant, but fundamentally different features.
Therefore, even models with identical test performance could have, in theory, picked up completely different cues to solve the same problem, hence yielding deviant explanations per model.
When training models using different training strategies, it cannot be avoided to obtain models with deviating classification strategies.
This is also clearly reflected in the naturally lower model performance of \textit{DP}-based models.

Further limitations are related to the evaluation of the explanation's quality through quantitative metrics.
As already mentioned, quantitative quality metrics for XAI are simply mathematical approximations of factors that could account for human interpretability or test assumptions of fidelity that should be satisfied by good explanations.
Many such metrics still have inherent limitations like \textit{AOPC}, \textit{Sensitivity}, and \textit{Infidelity}, introducing out-of-distribution samples through the perturbation of samples.
\textit{Ground Truth Concordance} assumes that, for each sample, there is exclusively one single decision path, and therefore a ground truth explanation.
To approach this assumption, a synthetic dataset was utilized that allowed the construction of ground truth segmentation maps, highlighting all decision-relevant shapes.
For somewhat complicated problems, explanations are always redundant as are the corresponding human problem definitions.
The human-made logic behind the dataset postulates that a set of pre-defined shapes (i. e., star or triangle) need to be present to associate a sample with a class.
However, instead of selecting the entirety of a shape, networks could also simply define triangles by the angle of their apices.
This way, the explanation would not need to cover the complete shape, but only an arbitrarily small area around a single apex.

\section{Conclusion}
\label{sec:conclusion}

Both eXplainable AI and privacy-preserving machine learning constitute pivotal technologies for the safe translation of state-of-the-art AI algorithms into everyday applications.
It is particularly important to get an early understanding of the effect of private training on XAI, to actively develop countermeasures, and avoid blind interpretations of explanations.
This work showed that, although the exact effect on explanations depends on a multitude of factors including the privacy technique, dataset, model architecture, and XAI method, some overall trends can be identified.
It has been found that \textit{Differential Privacy}, on average, decreases both the \textit{Interpretability} and \textit{Fidelity} of heatmaps.
However, \textit{Federated Learning} was found to moderate both effects when used in combination.
When used, alone, \textit{FedAVG} was even sometimes found to improve the \textit{Interpretability} of attribution maps by generating more continuous heatmaps.
The results suggest to consider \textit{Federated Learning} before \textit{Differential Privacy}, where appropriate.
Moreover, it is recommended always to choose \textit{Differential Private Federated Learning} as well as perturbation-based XAI methods, if an application requires both privacy and explainability.
As the first work to investigate the impact of privacy on XAI, this study opens up a series of interesting follow-up questions, including the in-depth analysis of the trade-off between privacy and interpretability under different privacy constraints, and the impact of using privacy techniques beyond \textit{DP} and \textit{FedAVG}.
Moreover, the field would benefit from a deeper analysis of the effect of privacy on the interpretation by human users through application-grounded and human-grounded evaluation methods.
The union of PPML and XAI solves one of the remaining regulatory and safety-critical hurdles, freeing the way for innovative and high-performing AI-based applications that can bring significant advancements in crucial domains of our everyday lives.

\bibliography{main}
\bibliographystyle{IEEEtran}

\clearpage

\begin{IEEEbiography}[{\includegraphics[width=1in,height=1.25in,clip,keepaspectratio]{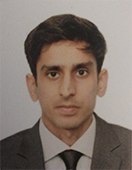}}]{Saifullah} received the B.S. degree in mechanical engineering and the M.S. degree in robotics and intelligent machine engineering from the National University of Sciences and Technology (NUST), Pakistan. He is currently pursuing his Ph.D. at the University of Kaiserslautern and is working as a researcher at the German Research
Center for Artificial Intelligence (DFKI GmbH) under the supervision of Prof. Dr. Prof. H. C.
Andreas Dengel. His research interests include document understanding and analysis, explainability and robustness of deep learning models, and privacy preservation in deep learning.
\end{IEEEbiography}

\begin{IEEEbiography}[{\includegraphics[width=1in,height=1.25in,clip,keepaspectratio]{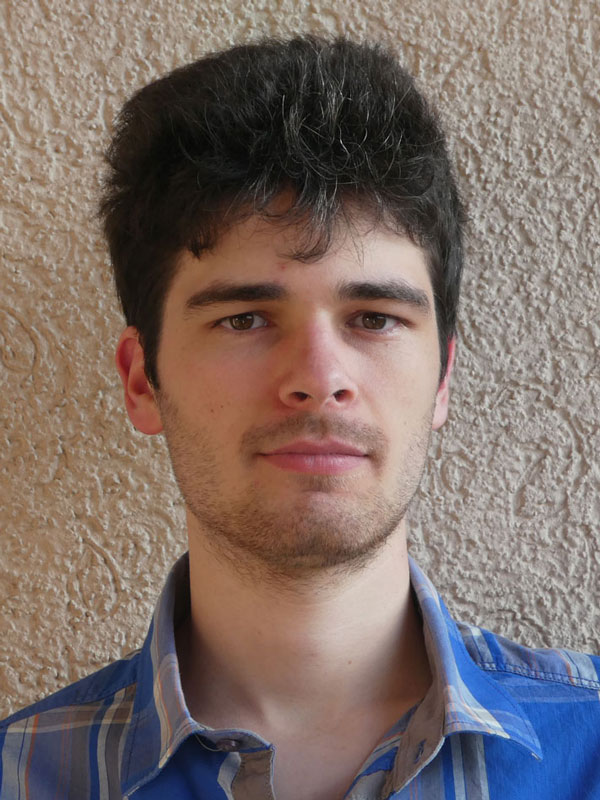}}]{Dominique Mercier} received his Master degree in computer science from the Technische Universitaet Kaiserslautern, Germany in 2018. The topic of his Master's thesis was 'Towards Understanding Deep Networks for Time Series Analysis'. Currently, he is pursuing his Ph.D. at the German Research Center for Artificial Intelligence (DFKI GmbH) under the supervision of Prof. Dr. Prof. h.c. Andreas Dengel. His areas of interest include the interpretability of deep learning methods, time series analysis, and document analysis. His work includes the development of novel interpretability methods for deep neural networks for time series analysis. Furthermore, he actively working in the NLP domain with a focus on citation and community management.
\end{IEEEbiography}

\begin{IEEEbiography}[{\includegraphics[width=1in,height=1.25in,clip,keepaspectratio]{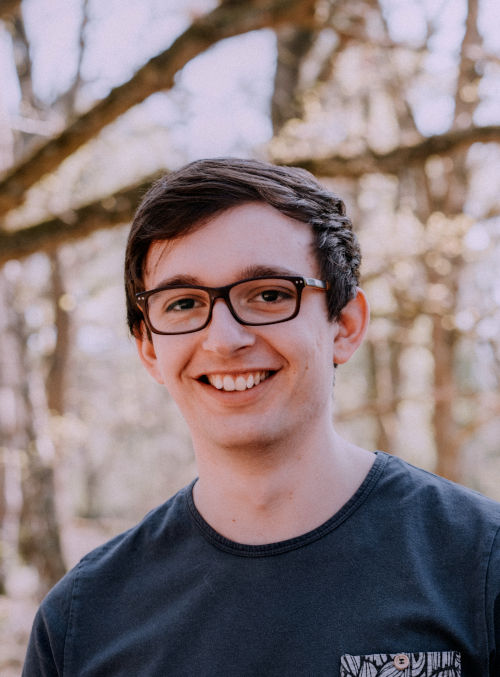}}]{Adriano Lucieri} completed his BE in Mechatronic Engineering from Duale Hochschule Baden-Württemberg (DHBW) Mannheim and MS in Mechatronic Systems Engineering from Hochschule Pforzheim in Germany. He is presently pursuing a Ph.D. from Technische Universität Kaiserslautern (TUK), Germany, and is also working as Research Assistant at Deutsches Forschungszentrum für Künstliche Intelligenz GmbH (DFKI). His research focus lies on improving the explainability and transparency of Computer-Aided Diagnosis (CAD) systems based on Deep Learning for medical image analysis. His work includes a concept-based explanation of skin lesion classifiers as well as the localization of concept regions in input images. 
\end{IEEEbiography}

\begin{IEEEbiography}[{\includegraphics[width=1in,height=1.25in,clip,keepaspectratio]{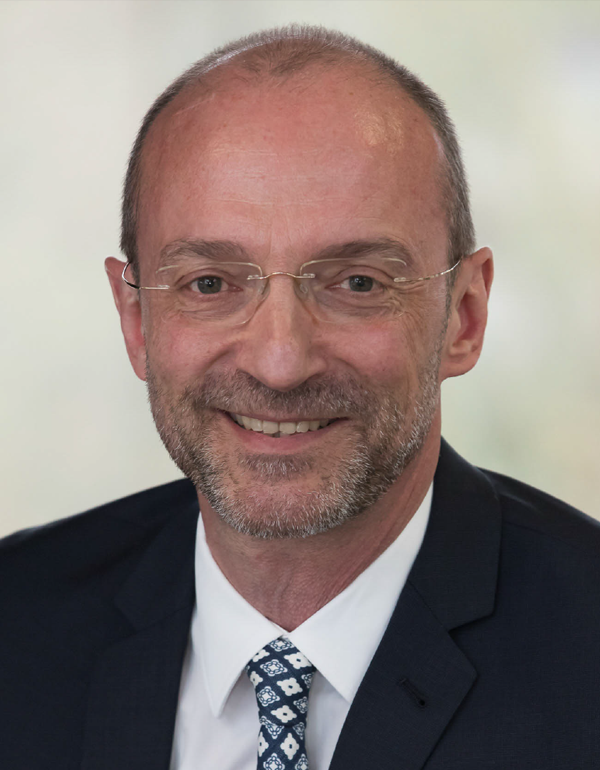}}]{Andreas Dengel} is Scientific Director at DFKI GmbH in Kaiserslautern. In 1993, he became Professor in Computer Science at TUK where he holds the chair Knowledge-Based Systems. Since 2009 he is appointed Professor (Kyakuin) in the Department of Computer Science and Information Systems at Osaka Prefecture University. He received his Diploma in CS from TUK and his Ph.D. from the University of Stuttgart. He also worked at IBM, Siemens, and Xerox Parc. Andreas is a member of several international advisory boards, has chaired major international conferences, and founded several successful start-up companies. He is a co-editor of international computer science journals and has written or edited 12 books. He is the author of more than 300 peer-reviewed scientific publications and supervised more than 170 Ph.D. and master theses. Andreas is an IAPR Fellow and received many prominent international awards. His main scientific emphasis is in the areas of Pattern Recognition, Document Understanding, Information Retrieval, Multimedia Mining, Semantic Technologies, and Social Media.
\end{IEEEbiography}

\begin{IEEEbiography}[{\includegraphics[width=1in,height=1.25in,clip,keepaspectratio]{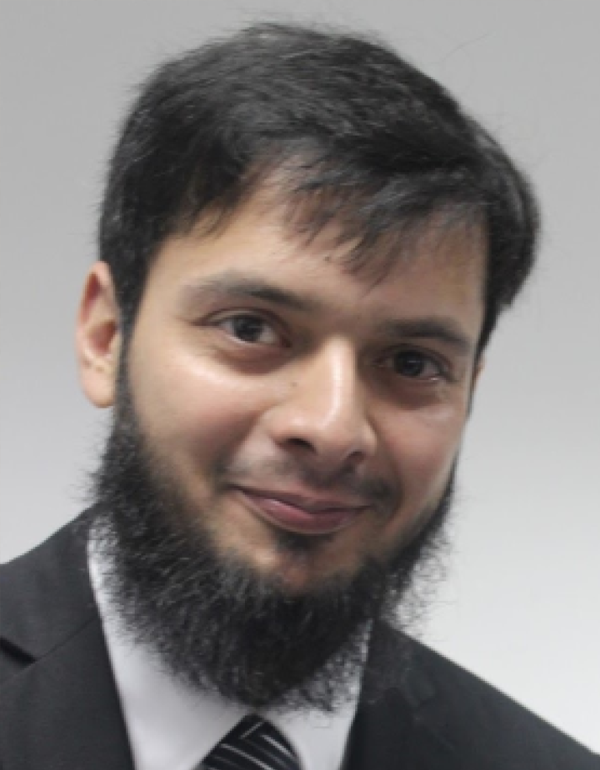}}]{Sheraz Ahmed} is Senior Researcher at DFKI GmbH in Kaiserslautern, where he is leading the area of Time Series Analysis. He received his MS and Ph.D. degrees in Computer Science from TUK, Germany under the supervision of Prof. Dr. Prof. h.c. Andreas Dengel and Prof. Dr. habil. Marcus Liwicki. His Ph.D. topic is Generic Methods for Information Segmentation in Document Images. Over the last few years, he has primarily worked on the development of various systems for information segmentation in document images. His research interests include document understanding, generic segmentation framework for documents, gesture recognition, pattern recognition, data mining, anomaly detection, and natural language processing. He has more than 30 publications on the said and related topics including three journal papers and two book chapters. He is a frequent reviewer of various journals and conferences including Pattern Recognition Letters, Neural Computing and Applications, IJDAR, ICDAR, ICFHR, and DAS.
\end{IEEEbiography}

\EOD

\end{document}